%% file: paper.tex
\documentclass[journal]{IEEEtran}
\usepackage{array}
\usepackage{multirow}
\usepackage{tikz,pgfplots}
\pgfplotsset{compat=1.15}
\usepgfplotslibrary{statistics}
\usepackage{xspace}
\usepackage{xcolor} 
\usepackage{booktabs}
\usepackage{hyperref}
\usepackage{footnote}
\usepackage{numprint}
\newcommand{\etal}{\textit{et\,al.}\xspace}
\newcommand{\eg}{e.\,g.\xspace}

\newcolumntype{R}[1]{>{\raggedleft\arraybackslash\hspace{0pt}}p{#1}}
\newcolumntype{L}[1]{>{\raggedright\arraybackslash\hspace{0pt}}p{#1}}
\newcolumntype{C}[1]{>{\centering\hspace{0pt}}p{#1}}
\newlength \figureheight
\newlength \figurewidth

\usepackage{cite}

\ifCLASSINFOpdf
\else
\fi
\usepackage{CJKutf8}

\usepackage{amsmath}
\ifCLASSOPTIONcompsoc
  \usepackage[caption=false,font=normalsize,labelfont=sf,textfont=sf]{subfig}
\else
  \usepackage[caption=false,font=footnotesize]{subfig}
\fi
\begin{document}
%
\title{Benchmarking Probabilistic Deep Learning Methods for License Plate Recognition
 }
%
%
%
\author{Franziska~Schirrmacher, Benedikt~Lorch, Anatol~Maier, and~Christian~Riess,~\IEEEmembership{Senior~Member,~IEEE}
\thanks{F.~Schirrmacher, B.~Lorch, A.~Maier, and C.~Riess are with the IT Security Infrastructures Lab, Computer Science, Univ. of Erlangen-N\"urnberg.}
\thanks{We gratefully acknowledge support of our work by the German Research Foundation (146371743/TRR 89), the German Research Foundation, GRK Cybercrime (393541319/GRK2475/1-2019), anf the German Federal Ministry of Education and Research (BMBF) under grant number 13N15319.}}%

\maketitle

\begin{abstract}




Learning-based algorithms for automated license plate recognition implicitly
assume that the training and test data are well aligned.
However, this may not be the case under extreme environmental conditions, or in
forensic applications where the system cannot be trained for a specific
acquisition device.
%
%
Predictions on such out-of-distribution images have an increased chance of failing. But this failure case is oftentimes hard to recognize for a human
operator or an automated system.
%

Hence, in this work we propose to model the prediction uncertainty
for license plate recognition explicitly. Such an uncertainty measure allows to detect
false predictions, indicating an analyst when not to trust the result of the
automated license plate recognition.

In this paper, we compare three methods for uncertainty quantification on two
architectures. 
The experiments on synthetic noisy or blurred low-resolution images show that
the predictive uncertainty reliably finds wrong predictions. We also show that
a multi-task combination of classification and super-resolution improves the
recognition performance by 109\% and the detection of wrong predictions by
29\%.

\end{abstract}


%

%
\IEEEpeerreviewmaketitle

\section{Introduction}
\label{sec:introduction}
%
%
%
%
\IEEEPARstart{L}{icense} plate recognition (LPR) is the task of detecting and deciphering the license plate number of a vehicle in an image. Performing this task in an automated procedure is of particular interest in traffic control, self-driving cars, or traffic surveillance~\cite{Silva2018,Zhang2020}. The wide range of possible applications entails challenges for the license plate recognition methods. 

In controlled environments, like toll monitoring, the images are captured from a frontal view and are of high quality. For these kinds of images, performing the task in an automated procedure is feasible with neural networks~\cite{Silva2018}. However, the acquisition scenario can vary in the wild, leading to rotated, blurred, or low-lighting images.~\cite{Silva2018}. A fast-moving vehicle, for example, can cause strong motion blur in the image. Even worse conditions, such as low-cost cameras, can lead to severely degraded images on which the license plate number is barely visible~\cite{Kaiser2021}. In such unconstrained scenarios in the wild, out-of-distribution examples pose a challenge to the neural networks. When the test images differ too much from the training distribution, neural networks are prone to silent failures. Since the exact acquisition setup is unknown, the training data needs to cover different combinations of camera models, environmental factors, and image degradation types. However, cost and effort to cover all of these conditions put the feasibility of a truly ``complete'' dataset into question.

Modeling the uncertainty of neural networks regarding its prediction has become an increasingly important area of research because of similar challenges in computer vision and image forensics~\cite{Kendall2017,Snoek2019,Maier2020}. In addition to the prediction of the neural network, a confidence estimate, called predictive uncertainty, gives a clue whether to trust the prediction. Ovadia \etal~\cite{Ovadia2019} group these approaches under the name probabilistic deep learning. Possible techniques to gather confidence estimates, among others, are Bayesian neural networks (BNN)~\cite{Kendall2017}, deep ensembles~\cite{Lakshminarayanan2017}, or Monte Carlo dropout~\cite{Gal2016}. Each of these techniques are explained in detail in Sec.~\ref{sec:relatedWork} and Sec.~\ref{sec:methods}.

\begin{figure}
  \centering
  \includegraphics[width=0.49\textwidth]{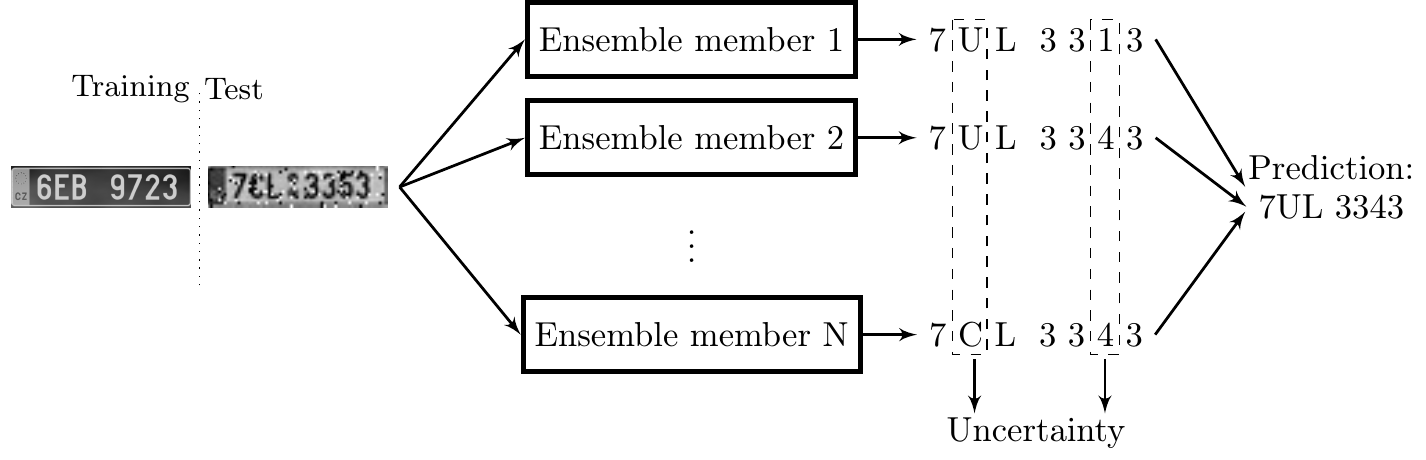}
  \caption{ Graphical illustration of quantifying uncertainty  with a deep ensemble for license plate recognition. Each ensemble member receives the same input image shown on the left. In case of training-test mismatch, the predictions of the ensemble members vary. The difference in the predictions encodes the uncertainty of the ensemble regarding the predicted characters. }
  \label{fig:teaser}
\end{figure}

Such confidence estimates are worth investigating in the context of license plate recognition. {We consider these confidence estimates to strengthen two applications: automatic license plate recognition (ALPR) and forensic license plate recognition (FLPR). To the best of our knowledge, applying probabilistic deep learning techniques to license plate recognition has not been explored yet. Therefore, we investigate the possible use cases of these techniques for license plate recognition. We go beyond merely comparing the methods and provide additional insights into the uncertainty measures, \eg the detection of false prediction. 
}

In ALPR systems, all recognition steps are performed automatically without human interaction and the license plate number is still readable~\cite{Silva2018}. Therefore, the confidence estimate can provide an incentive for a human to verify the prediction. 

In FLPR, police investigators receive images from various camera models and acquisition setups, \eg, surveillance cameras of petrol stations. The images are analyzed manually, so human-readable license plates of reduced quality are not a problem. More problematic are license plates that are indecipherable due to poor image quality. Images of such low quality require reliable license plate recognition since the police investigator is not able to verify the prediction of the network. The confidence estimate can indicate a potentially falsely predicted character within the license plate number.


In this paper, we examine three probabilistic deep learning techniques that provide an additional confidence measure: Deep ensembles~\cite{Lakshminarayanan2017}, BatchEnsemble~\cite{Wen2020}, and Monte Carlo dropout~\cite{Gal2016}. The methods are selected because they offer a more scalable alternative to BNNs, which are typically associated with predictive uncertainty. Figure~\ref{fig:teaser} graphically illustrates the functionality of probabilistic deep learning by means of deep ensembles. The input image is processed by multiple ensemble members whose predictions vary when the input distribution differs from the training distribution. The ensemble members generate the uncertainties that can then provide helpful feedback to the analyst.

A convolutional neural network (CNN) proposed by Lorch \etal~\cite{Lorch2019} and a multi-task framework inspired by SR$^2$~\cite{Schirrmacher2020} serve as backbone for the probabilistic deep learning techniques. The CNN~\cite{Lorch2019} has been used several times for license plate recognition and therefore represents a good baseline~\cite{Rossi2021,Kaiser2021}. {So far, work on multi-task learning for license plate recognition proposes a sequential ordering of first image processing and then LPR~\cite{8462282, Lee_2019_ICCV, Rossi2021}. However, the sequential ordering is prone to error propagation and less robust when faced with out-of-distribution data~\cite{Schirrmacher2020}. For this reason, we propose a parallel arrangement of license plate recognition and super-resolution in the SR$^2$ framework. We can show that super-resolution introduces an inductive bias that benefits license plate recognition. Additionally, the SR$^2$ framework provides better generalization and extraction of relevant features than individual tasks only~\cite{Caruana1993, Caruana1997, Ruder2017}. As a result, the character recognition improves in the SR$^2$ framework. Super-resolution, unlike classification, is susceptible to noise or other types of degradation which are not present in the training data~\cite{Schirrmacher2020,Villar2021}. Thus, super-resolution improves the expressiveness of predictive uncertainty. }

For both architectures, high-quality images of synthetic Czech license plates are the basis of the training set. We utilize the data generation pipeline proposed by Kaiser \etal~\cite{Kaiser2021}. To mimic out-of-distribution samples, we corrupt the test set with different forms of unseen image degradations, namely noise or blur. {In our experience, }these degradations {most frequently} appear in real world data and are usually covered in license plate recognition datasets. {It should be noted that modeling the complex patterns of degradations and their magnitude in the real world is challenging. Therefore, we can assume that out-of-distribution data will occur in real world scenarios, since covering all types and strengths of degradations in the training data is infeasible.}



The experiments show that we can reliably detect wrong predictions by looking at the predictive uncertainty. Compared to the baseline, the SR$^2$ framework improves the character accuracy and the expressiveness of the predictive uncertainty.
Also, the combination of the license plate recognition CNN with MC-dropout achieves competitive results compared to related work, with the additional benefit of the predictive uncertainty\footnote{The source code to our methods can be downloaded at: \url{https://github.com/franziska-schirrmacher/LPR-uncertainty}.}.


The contributions of the paper are three-fold:
\begin{enumerate}
  \item We improve the \emph{reliability} of license plate recognition {by utilizing three probabilistic deep learning techniques: deep ensembles, BatchEnsemble, and Monte Carlo dropout. To the best of our knowledge, this is the first work to explore probabilistic deep learning for automatic license plate recognition. We investigate the opportunities of this approach, with particular focus on the automated detection of misclassifications.}
  \item The methods are tested on out-of-distribution data to compare the expressiveness of the predictive uncertainty for license plate recognition. To this end, the test images contain different forms of degradation, specifically noise and blur. We show that the deep ensemble and MC-dropout achieve the best results. 
  \item  We improve the predictive uncertainty within a multi-task learning framework that combines super-resolution and license plate recognition. The experiments show that the combination outperforms license plate recognition only in terms of accuracy and predictive uncertainty.
\end{enumerate}

The paper is organized as follows: Section~\ref{sec:relatedWork} provides an overview on related work for license plate recognition and predictive uncertainty. The probabilistic deep learning methods are described in Sec.~\ref{sec:methods} along with the license plate recognition and the super-resolution network. The experiments and the used dataset are reported in Sec.~\ref{sec:experiments}. Section~\ref{sec:conclusion} concludes the paper.

\section{Related Work}
\label{sec:relatedWork}

This section comprises the related work of license plate recognition and predictive uncertainty. 

\subsection{License Plate Recognition}

License plate recognition involves multiple steps which are performed in sequence~\cite{Shashirangana2020}. First, the car and the license plate are detected. Then, after extracting the region-of-interest, the license plate number is classified. 
To extract the license plate position within the image, global information, texture, or color features can be used~\cite{Du2012}. A classifier predicts the license plate number in the extracted image region~\cite{Anagnostopoulos2008,Du2012}. First, the characters are segmented using prior knowledge, projection profiles, or character contours. The subsequent recognition step utilizes templates~\cite{Du2012}, strokes~\cite{Wen2011}, or local binary patterns~\cite{Liu2010}.

Due to the success of neural networks in object detection~\cite{Redmon2016,Ren2016,Tan2020} and classification~\cite{Krizhevsky2012,He2016, Tan2019}, state-of-the-art license plate recognition methods utilize a sequence of neural networks to combine these two tasks. In some cases, the networks may fuse these steps into a single end-to-end pipeline~\cite{Li2018}. 

For license plate detection, YOLO~\cite{Redmon2016} is a popular choice~\cite{Zhang2020,Silva2018,Silva2021}. Zhang \etal~\cite{Zhang2019} use a Mask R-CNN~\cite{He2017}. Qiao \etal~\cite{Qiao2020} replace license plate detection with a position-aware mask attention module that directly detects characters in the image. Li \etal~\cite{Li2018} propose a parallel end-to-end approach of license plate detection and recognition. Both tasks share convolutional features and are then split into two branches.

For license plate recognition, there are two main neural network architectures currently adopted in the research. One design consists of only convolutional layers~\cite{Silva2018,Agarwal2017,Kaiser2021}. Another design utilizes a combination of convolutional and recurrent layers~\cite{Shi2016,Shivakumara2018,Suvarnam2019,Zhang2019}. The feature sequence generated by the recurrent layers is transcribed into a label using connectionist temporal classification~\cite{Graves2006}. 

A particular challenge for all recognition systems are low quality images. In police investigations, for example, low-quality images impede the investigation. Low-resolution and compression, to name the most limiting factors, prohibit reading the license plate number by looking at the image.

There are two different approaches to improve the recognition performance on low quality images: some work has been done in license plate recognition in combination with image processing. Others train a neural network on very low-quality images.

A combination of image denoising and license plate recognition is proposed by Rossi \etal~\cite{Rossi2021}. The first convolutional neural network removes the noise in the image. The denoised and noisy images are both processed by the second CNN to predict the license plate number. Seibel \etal~\cite{Hilario2017} combine multi-frame super-resolution and two optical character recognition (OCR) systems in a sequential framework for low-quality surveillance cameras. Schirrmacher \etal~\cite{Schirrmacher2020} showed that the combination of super-resolution and character recognition, called SR$^2$, achieves superior performance when performed in parallel. The parallel arrangement mitigates the issue of error propagation and makes the super-resolution more robust to unseen noise.

The training dataset in~\cite{Agarwal2017} contains low-resolution and noisy US license plates. With this dataset, a CNN can recognize two sets of three characters each of license plates in images with a reduced quality. Lorch \etal~\cite{Lorch2019} extend their CNN to seven separate outputs, one for each character. A null character allows for the recognition of license plates of varying lengths. Kaiser \etal~\cite{Kaiser2021} additionally investigate the influence of compression on the recognition rate of synthetic Czech license plates.

{ For both ALPR and FLPR, detection of license plates from different countries is challenging besides varying image degradations in unconstrained settings~\cite{Henry2020}. One challenge is the availability of public datasets. This allows related work to test their methods at least on some countries~\cite{Li2019}. Another challenge is the layout of the license plate that can greatly vary between countries. As previous work has shown~\cite{Kaiser2021}, the neural network learns which character is possible at which position. This may lead to a training-test mismatch when only limited training datasets are available~\cite{Henry2020}. We see great potential in probabilistic deep learning for multi-national license plate recognition. The probabilistic network can be trained on the existing datasets to cover license plates some countries. Predictions of license plates whoes layout varys greatly from the license plates in the training dataset are then marked as uncertain by the network.}

\subsection{Predictive Uncertainty}

Out-of-distribution samples most likely appear in real-world applications. Common feed forward neural networks guess their prediction on out-of-distribution data. Quantifying predictive uncertainty allows specifying the reliability of a neural network's prediction. Predictive uncertainty is often split into aleatoric and epistemic uncertainty. Epistemic uncertainty expresses the uncertainty of the model and can be decreased by adding more training data. Aleatoric uncertainty captures the uncertainty regarding the data, \eg due to noise in the observations or labels~\cite{DerKiureghian2009, Kendall2017}. In this work, we do not differentiate between aleatoric and epistemic uncertainty and only estimate predictive uncertainty.

Some works directly estimate predictive uncertainty based on the network's prediction, \eg, the softmax output~\cite{Hendrycks2017}. Guo \etal~\cite{Guo2017}, for example, propose to rescale the output of the neural network in a post-processing step. However, it has been shown that these softmax statistics can be misleading~\cite{Maier2020}.

A principled approach to obtain predictive uncertainties is via Bayesian modeling~\cite{Hinton1993,Graves2011,Blundell2015,Snoek2019}. To this end, Bayesian neural networks (BNN) learn a distribution over possible weights. The predictive distribution is obtained by marginalizing over the weight distribution. In practice, however, BNNs require restrictive approximations of the weight distributions or expensive numerical sampling. Due to these difficulties, more scalable alternatives to BNNs have been developed. 

The most straight-forward alternative to estimate predictive uncertainty are deep ensembles~\cite{Lakshminarayanan2017}. A deep ensemble comprises multiple neural networks which are trained on the same task. Wen \etal propose BatchEnsemble~\cite{Wen2020}, an ensemble technique that requires significantly fewer computations and memory than deep ensembles. Gal \etal~\cite{Gal2016} introduce Monte Carlo dropout (MC-dropout), which allows obtaining predictive uncertainty by applying dropout also during testing. In this paper, we consider deep ensemble, BatchEnsemble, and MC-dropout for our experiments.

\section{Methods}
\label{sec:methods}

This section provides a concise description of the methods employed in the experiments. The first part gives an overview of the two neural network architectures that serve as the base for the probabilistic deep learning methods. The second part explains the three probabilistic deep learning methods deep ensemble, BatchEnsemble, and MC-dropout.

\subsection{Backbone Neural Network Architectures for License Plate Recognition}

We evaluate the efficacy of the probabilistic deep learning methods with two neural network architectures as a backbone. The first backbone is a license plate recognition CNN \cite{Lorch2019, Kaiser2021}. The second backbone employs license plate recognition and super-resolution in the multi-task learning framework SR$^2$~\cite{Schirrmacher2020}. 

W adapt the following implementation details for both backbones: all neural network architectures use ReLU as an activation function. Additionally, batch normalization~\cite{Ioffe2015} is performed after each trainable layer. As a result, the following order is used in all architectures presented in the paper: trainable layer - batch normalization - ReLU.

\subsubsection{License Plate Recognition CNN}
The license plate recognition CNN consists of convolutional layers, max pooling layers, and fully-connected layers. All convolutional layers have a receptive field of $3\times3$. The max pooling layers have a pool size of $2\times2$. In contrast to~\cite{Kaiser2021}, our proposed license plate recognition CNN contains additional batch normalization layers to stabilize training. 

The architecture is structured as follows. After the input, there are three sequences of two convolutional layers followed by a max pooling layer with 64 filter kernels, 128 filter kernels, and 256 filter kernels, respectively. Then, there are two blocks, each with a convolutional layer with 512 filter kernels followed by max pooling. Then, the features maps are flattened followed by two fully-connected layers with 1024 and 2048 nodes. Finally, the CNN has seven fully-connected output layers with 37 nodes each. Here, softmax replaces the ReLU activation, and batch normalization is omitted. 

\subsubsection{SR$^2$}

SR$^2$ consists of shared layers followed by a split into two branches, one for super-resolution and the other for license plate recognition. We use FSRCNN~\cite{Dong2016} for super-resolution and the baseline CNN~\cite{Lorch2019} for the license plate recognition. FSRCNN was selected because its first layer is similar to that of the license plate recognition CNN (LPR CNN). Therefore, the shared layers in the SR$^2$ framework do not differ from the original layers of the individual CNNs. 

FSRCNN consists of five steps. First, features are extracted using a convolutional layer with 56 filter kernels and a receptive field of $5 \times 5$. Then, the feature maps are shrunk using 12 filter kernels with a receptive field of $1 \times 1$. Afterward, four convolutional layers with 12 filter kernels each and a receptive field of $3 \times 3$ perform a mapping. Next, one convolutional layer with 56 filter kernels and a receptive field of $1 \times 1$ expands the features maps. The last step is a convolutional layer with 192 filter kernels and a receptive field of $9 \times 9$ followed by a pixel shuffling.

In line with~\cite{Schirrmacher2020}, we propose a parallel arrangement of the license plate recognition CNN and the super-resolution FSRCNN. The tasks share the first convolutional layer of FSRCNN and then split into two branches in our setup. The loss function is the weighted sum of the individual loss functions. We choose $w_{lpr} = 20$ and $w_{sr} = 1$ as weights for the loss of the license plate recognition and super-resolution, respectively.

The addition of a super-resolution branch to the classification network boosts the predictive uncertainty of the classification. Super-resolution is particularly useful in studying predictive uncertainty since it is more sensitive to unseen degradations~\cite{Villar2021}. Therefore, smaller degradations potentially lead to higher predictive uncertainty. In general, multi-task learning helps to generalize and acts as a regularizer. Thus, the license plate recognition performance also benefits from the additional task. The multi-task framework can better identify relevant features when both tasks extract the same features, such as the edges of the characters~\cite{Caruana1993, Caruana1997, Ruder2017}.

\subsection{Uncertainty Quantification Methods}

This paper compares three different probabilistic deep learning methods in the context of license plate recognition. These methods are Monte Carlo dropout~\cite{Gal2016}, deep ensembles~\cite{Lakshminarayanan2017}, and BatchEnsemble~\cite{Wen2020}. Dropout is a common technique to regularize neural networks. The work by Lorch \etal~\cite{Lorch2019}, for example, uses dropout in their CNN. Thus, models trained with dropout can benefit from our findings without the need for re-training. Existing methods that do not utilize dropout during training might consider using their pipeline to train multiple models to get a deep ensemble. As deep ensembles require high computational power and memory, we additionally explore the efficacy of BatchEnsemble on the task of license plate recognition. The remainder of this Section presents details on the configuration of these three approaches.

 

Dropout~\cite{Srivastava2014} is a well-known regularization technique in the area of deep learning. Typically, dropout is applied only during training. However, Gal \etal~\cite{Gal2016} propose to use dropout not only during training but also during testing, called Monte Carlo dropout (MC-dropout). The inference step is performed multiple times. They show that the obtained variance between the predictions "minimizes the Kullback-Leibler divergence between an approximate distribution and the posteriori distribution of a Gaussian process"~\cite{Gal2016}. Thus the obtained predictive uncertainty gives a valid statement. Inference runs are performed multiple times with the same data to quantify the predictive uncertainty. Due to the random dropout of nodes, each inference run gives slightly different results. The number of trainable parameters is adapted to ensure that higher dropout rates do not reduce the representational power of the model. Therefore, the number of filters in each convolutional layer and the number of nodes in the fully-connected layers are increased by a factor of $\sqrt{1 / (1 - r)}$, where $r$ denotes the dropout rate. Thus, the CNN with dropout rate $r = 0.5$ has twice the trainable parameters compared to the CNN without dropout. Without the square in the factor, the parameter would quadruple since the number of trainable weight scales with the input size and the current layer.

Arguably, deep ensembles are the most straight-forward approach to estimate predictive uncertainty~\cite{Lakshminarayanan2017}. A deep ensemble comprises multiple neural networks which are trained on the same task. Due to random initialization of the weights and random data shuffling, each model ends up in a different local minimum with high probability~\cite{Fort2019}. Thus, the trained parameters are different in each model. During testing, the difference between the models' predictions expresses predictive uncertainty. However, the training of deep ensembles is time-consuming, and the deep ensemble requires much memory. Since each ensemble member has a similar behavior but makes different errors, uncertainty can be quantified well. The performance of the ensemble members is a lower bound for the overall performance of the ensemble~\cite{Krogh1994}. However, training is time-consuming. Additionally, the deep ensemble requires much memory since each ensemble member has to be trained and stored individually. 

BatchEnsemble~\cite{Wen2020} is an ensemble-based method that requires significantly fewer computations and memory than deep ensembles. The weight matrix of each ensemble member is constructed from two matrices. The first matrix is a full-rank weight matrix shared across all ensemble members. The second matrix is a rank-one matrix that is unique for each ensemble member. Using the Hadamard product between these shared and individual weights results in differing weights for each ensemble member. Therefore, a BatchEnsemble is trained and tested within one run by replicating the test data according to the number of ensemble members. Additionally, BatchEnsemble generates with its rank-1 matrices only a small memory overhead. For this paper, we use the code provided by the authors.

Each of the methods mentioned above provides multiple predictions for the same input data. To estimate predictive uncertainty, we compute the standard deviation of the predictions at every position of the output vector.

\section{Experiments}
\label{sec:experiments}

This section presents four different experiments on predictive uncertainty in license plate recognition. First, we show which of the probabilistic deep learning methods is best suited for license plate recognition. For this, we use the license plate recognition CNN as a backbone. To better understand the predictive uncertainty obtained by MC-dropout, we investigate the influence of the dropout rate and the number of inference runs in an ablation study. After that, we show the benefit of the inductive bias introduced by the super-resolution in the SR$^2$ backbone. The last experiment demonstrates the competitive performance of MC-dropout on a real world dataset. To this end, we compare MC-dropout with the LPR CNN as a backbone to related work.

\subsection{Experimental Setup}

\begin{figure}[!tbp]
	\centering
	\subfloat[High-resolution image]{\includegraphics[width=0.2\textwidth]{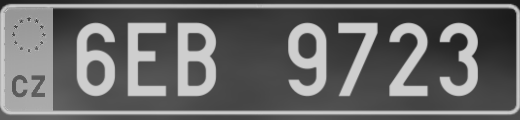}} \
  \subfloat[Low-resolution image \label{fig:example_lr}]{\includegraphics[width=0.2\textwidth]{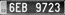}}

	\subfloat[Gaussian $\sigma=0.1$ \label{fig:example_g1}]{\includegraphics[width=0.15\textwidth]{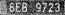}} \
	\subfloat[Gaussian $\sigma=0.2$]{\includegraphics[width=0.15\textwidth]{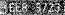}} \
	\subfloat[Gaussian $\sigma=0.3$]{\includegraphics[width=0.15\textwidth]{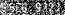}} \
	
	\subfloat[S\&P $p=0.1$ \label{fig:example_sp_01}]{\includegraphics[width=0.15\textwidth]{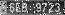}} \
	\subfloat[S\&P $p=0.2$]{\includegraphics[width=0.15\textwidth]{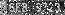}} \
	\subfloat[S\&P $p=0.3$]{\includegraphics[width=0.15\textwidth]{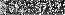}} \
	
	\subfloat[Horizontal $k=3$]{\includegraphics[width=0.15\textwidth]{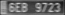}} \
	\subfloat[Horizontal $k=5$]{\includegraphics[width=0.15\textwidth]{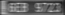}} \
	\subfloat[Horizontal $k=7$]{\includegraphics[width=0.15\textwidth]{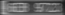}} \
	
	\subfloat[Vertical $k=3$]{\includegraphics[width=0.15\textwidth]{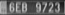}} \
	\subfloat[Vertical $k=5$]{\includegraphics[width=0.15\textwidth]{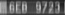}} \
	\subfloat[Vertical $k=7$ \label{fig:example_v3}]{\includegraphics[width=0.15\textwidth]{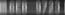}} \
	\caption{Example images from the synthetic DS-Hard dataset with different types and levels of degradation. The high-resolution image is first downsampled and then corrupted by additive Gaussian noise, salt \& pepper (S \& P) noise, horizontal blur, or vertical blur.}
  \label{fig:example}
\end{figure}

\paragraph{Datasets}
Synthetic data allows controlled experiments with different types of distortions. For this reason, a synthetic dataset is used to evaluate the efficacy of the probabilistic deep learning methods. The data generation pipeline proposed by Kaiser \etal~\cite{Kaiser2021} offers the possibility to generate grayscale Czech license plates. With the pipeline, high-resolution and low-resolution image pairs along with the corresponding license plate number are generated. First, the pipeline generates a high-resolution image with a size of $120 \times 520$ pixels. Second, a nearest-neighbor downsampling operation lowers the resolution of the image by a factor of 8 to $15 \times 64$ pixels. Finally, Gaussian noise with 20\,dB is added, and JPEG compression with a JPEG quality factor of 100 is applied. There is no rotation present in the image pairs. The low-resolution images are the input to the LPR CNN, and the license plate numbers are the corresponding labels. In the SR$^2$ setup, the high-resolution images serve as labels for the FSRCNN.

The dataset consists of a total of \numprint{110000} images, split 80k/20k/10k into training, validation, and test images.

The training and validation images are not further processed. The test set is additionally corrupted with noise or motion blur to mimic out-of-distribution data. The types of noise are additive Gaussian noise and salt \& pepper noise. We choose the standard deviation $\sigma \in {0.0001, ..., 0.4}$ for Gaussian noise according to visual appearance. A pixel is either set to full intensity (salt) or no intensity (pepper) with probability $p \in {0.0001, ..., 0.4}$\footnote{The precise values for both noise distributions, defined by $p$ and $\sigma$, are $10^{-1}\times{0.001, 0.01, 0.05, 0.1, 0.25,0.5, 0.75, 1,1.5, 2, 2.5, 3, 3.5, 4}$}. Salt and pepper pixels occur with equal probability. A horizontal and a vertical blur kernel with kernel size $k \in [3,5,7,11]$ smear the license plate characters. {In one experiment, we consider defocus blur by applying a Gaussian blur kernel with standard deviation $\sigma_f$ to smear the characters in all directions.}

In our experiments, we differentiate between two test datasets. The DS-Full dataset contains the entire test dataset. The DS-Hard dataset is a subset of the test dataset. Here, only strongly degraded images are considered namely with Gaussian noise with $\sigma \geq 0.1$, salt \& pepper noise with $p \geq 0.1$, and blur kernel $k = [3,5,7]$. These are are the strengths of degradation where the transition to an unreadable license plate number takes place. 

Figure~\ref{fig:example} visualizes example images for three different strengths of degradation. The first row shows the high-resolution image (left) and the low-resolution image without additional noise or blur (right). The image quality of the low-resolution image is equal to the image quality of the images in the training dataset. The second row shows the low-resolution images with additive Gaussian noise with $\sigma=0.1$ (left), $\sigma=0.2$ (middle), and $\sigma=0.3$ (right). The third row shows the low-resolution images with salt \& pepper noise with $p=0.1$ (left), $p=0.2$ (middle), and $p=0.3$ (right). The fourth row shows the low-resolution images with horizontal blur with kernel size $k=3$ (left), $k=5$ (middle), and $k=7$ (right). The last row shows the low-resolution images with vertical blur with kernel size $k=3$ (left), $k=5$ (middle), and $k=7$ (right). 

Additionally, we perform experiments on the CCPD base dataset~\cite{Xu2018}. The dataset contains labeled images with Chinese license plates captured from a city parking management company. Along with the bounding boxes of the license plates, the dataset provides the license plate numbers. 
Different weather conditions, rotation, and blur lead to images of varying image quality.
The provided training set is split into training and validation with \numprint{80000} and \numprint{20000} images accordingly. The proposed LPR CNN is tested on the base test set with \numprint{100000} images.

\paragraph{Evaluation protocol}

The license plate recognition CNN outputs seven vectors with 37 elements each. One vector represents one character in the license plate number. The position of the maximum element of each vector is the predicted character. Throughout the experiments, the characters in the license plate are considered individually except for the comparison on the CCPD dataset.

Each probabilistic deep learning method provides multiple predictions for one input image. Deep ensembles provide one prediction for each ensemble member. BatchEnsemble requires replication of the input according to the number of ensemble members. Then, the output of the BatchEnsemble is split into the individual predictions. MC-dropout performs a number of inference runs to obtain multiple predictions. For each method, the mean and the standard deviation of the predictions are computed. 
\begin{figure*}[!tbp]	
  \setlength \figurewidth{0.19\textwidth}
  \setlength \figureheight{1.0\figurewidth}
  \scriptsize
  
  \centering
  \includegraphics[width=0.6\textwidth]{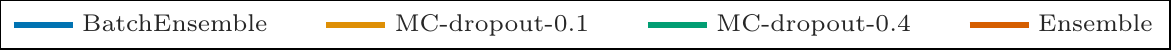}

\subfloat[Salt \& Pepper noise \label{fig:uncertainty_noise_accuracy_sp}]{\input{tex/noise-uncertainty-sp-aC-5}}  
  \subfloat[Gaussian noise \label{fig:uncertainty_noise_accuracy_gaussian}]{\input{tex/noise-uncertainty-gaussian-aC-5}}  
  \subfloat[Horizontal blur \label{fig:uncertainty_blur_accuracy_horizontal}]{\input{tex/blur-uncertainty-horizontal-aC-5}}
  \subfloat[Vertical blur \label{fig:uncertainty_blur_accuracy_vertical}]{\input{tex/blur-uncertainty-vertical-aC-5}}

  \caption{Accuracy on the DS-Full dataset. The mean character accuracy of the BatchEnsemble, dropout, and the deep ensemble on the synthetic test dataset. The low-resolution images are corrupted with horizontal and vertical blur with varying kernel size $k$. Moreover, Gaussian noise with standard deviation $\sigma$ or salt \& pepper noise with probability $p$ is added to the low-resolution images.}
  \label{fig:uncertainty_accuracy}
\end{figure*}


In the experiments, three different aspects of license plate recognition are considered. First, we evaluate the character recognition performance of the probabilistic deep learning techniques. Second, we test if the predictive uncertainty can be used to detect false predictions. Lastly, we investigate influencing factors that change the distribution of the predictive uncertainty values and thus the detection of false predictions.

We use the mean prediction to measure the accuracy of the character recognition. An accuracy of 1 means all characters in the test set are predicted correctly. The number of correctly predicted license plates is counted and divided by the total number of license plates in the dataset to compute the license plate accuracy.

The standard deviation of the predictions of each character represents the predictive uncertainty. A prediction can either be correct or false. Additionally, each prediction has been assigned a predictive uncertainty value. Using a threshold on the predictive uncertainty values, we can identify false predictions. We measure how well false predictions can be identified for a given threshold with the precision-recall curve. True positive is a predictive uncertainty value of a false prediction that is above the threshold. False positive is a predictive uncertainty value above the threshold that belongs to a correct prediction. False negative is a false prediction with a predictive uncertainty value below the threshold. We measure the area-under-the-curve (AUC) of the precision-recall curve with varying thresholds. A large AUC indicates well-separated predictive uncertainty values for correct and false predictions. To account for class imbalance, we perform random subsampling. The probabilistic deep learning techniques achieve a perfect character recognition performance on images with no degradation. With increasing strength of the degradation, a misclassification becomes more likely. For the neural networks, the strength of the degradation where the first misclassification happens can vary. Note that the computation of the AUC is possible only if there is a wrong classification. Therefore, we have a different number of AUC values for the different techniques. Thus, we compute the mean AUC of the precision-recall curve in all Tables only on the DS-Hard dataset. Also, these strongly degraded images are relevant for FLPA. Since the police investigator can not verify the network's prediction, he relies on the predictive uncertainty to identify false predictions.


The hyperparameter of the probabilistic deep learning techniques and the degradation strength change the distribution of the predictive uncertainty values. We quantize the parameters of the distribution of the predictive uncertainty values of correct and false predictions with the median and the interquartile range. The interquartile range is the difference between the 25th and 75th percentile and measures the spread of the distribution.

\paragraph{Training}

For deep ensemble, BatchEnsemble, and MC-dropout, we obtain five predictions. If not stated differently, five inference runs are performed with MC-dropout to ensure fairness to the other methods. The deep ensemble and MC-dropout are trained with a batch size of 32. The batch size of BatchEnsemble is raised to 160 by replicating the batch five times. For the deep ensemble and MC-dropout, we use He normal to initialize the trainable parameters. BatchEnsemble uses random initialization drawn from a normal distribution with a mean of 1.0 and a standard deviation of 0.5, as specified in the official implementation.

For each of the three probabilistic deep learning methods, either the license plate recognition CNN or SR$^2$ serves as a backbone. The loss is the averaged cross-entropy loss of each position. The mean absolute error loss is used for training the super-resolution network. The structural similarity index measure (SSIM) indicates potential overfitting when applied to the validation data.

Adam is used for optimization with the standard parameter $\beta_1 = 0.9$, $\beta_2 = 0.999$, and $\epsilon = 1e^{-7}$. The learning rate for dropout is set to $0.001$, while the deep ensemble and BatchEnsemble are trained with a learning rate of $0.00001$. We set the L2 kernel regularizer to $0.0001$ for dropout and $0.01$ for the deep ensemble and the BatchEnsemble. Each model is trainied for 55 epochs. Additionally, the learning rate is reduced during training when the validation loss stagnates for more than five epochs. The learning rate decay is set to $0.2$.

The models are trained using \textit{Tensorflow 2.4.1} and evaluated using \textit{scikit-learn 0.24.1}. The training ran on a NVIDIA GeForce RTX 2080 Ti GPU.
\begin{table}[!tbp]
	\centering
	\caption{Accuracy on the DS-Full dataset. Mean character recognition accuracy of the BatchEnsemble, dropout, and the deep ensemble. For each degradation type, the values represent the mean accuracy on all degradation levels.}
\begin{tabular}{lrrrr} 

  &  \multicolumn{2}{c}{noise} &  \multicolumn{2}{c}{blur}  \\
  \cmidrule(lr){2-3}\cmidrule(lr){4-5} 
  Method &  salt \& pepper &  Gaussian & horizontal & vertical\\  \hline
  BatchEnsemble & 0.691 & 0.832 & 0.230 & 0.280 \\
  MC-dropout-0.1 & 0.775 & 0.803 & 0.295 & 0.371 \\
  MC-dropout-0.4 &\textbf{0.844} & 0.801 &\textbf{0.524} &\textbf{0.550} \\
  Ensemble & 0.726 &\textbf{0.846} & 0.306 & 0.357 \\
   \end{tabular}
   \label{tab:uncertainty_accuracy}
\end{table}

{
\begin{table*}[!tbp]
	\centering
	\caption{Accuracy on the DS-Full dataset. Mean character recognition accuracy of the BatchEnsemble, dropout, and the deep ensemble on images impaired by two types of degradations. }
\begin{tabular}{lR{4.6cm}R{4.6cm}R{5cm}} 
   & 
   \includegraphics[width=4.6cm]{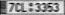} &
   \includegraphics[width=4.6cm]{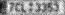} &
   \includegraphics[width=4.6cm]{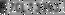} \\
  Method &  
  Gaussian ($\sigma$ = 0.01) horizontal (k = 3) &  
  Gaussian ($\sigma$ = 0.05)  defocus ($\sigma_f$ = 1.0)  & 
  Salt \& pepper (p = 0.1) defocus ($\sigma_f$ = 0.1)\\  \hline
  BatchEnsemble & 0.581 & 0.302 & 0.144  \\
  MC-dropout-0.1 & 0.811 & 0.366 & 0.251  \\
  MC-dropout-0.4 & \textbf{0.922} & \textbf{0.654} &\textbf{0.497}  \\
  Ensemble & 0.657 &0.244 & 0.139  \\
   \end{tabular}
   \label{tab:mixed}
\end{table*}}

\subsection{Comparison of Probabilistic Deep Learning Methods}
This section compares the accuracy and predictive uncertainty of BatchEnsemble, deep ensemble, and MC-dropout with the LPR CNN as a backbone. We evaluate different dropout rates $r$ for MC-dropout, denoted as MC-dropout-$r$. We test on out-of-distribution data to investigate the efficacy of the methods.

The reliability of the character recognition and the detection of misclassifications decreases with increasing strength of degradation. Both metrics, accuracy and AUC, are stable up to a certain point and then drop rapidly. This critical point depends on the type of degradation. Gaussian noise is less challenging than salt \& pepper noise and blur.

\subsubsection{Character recognition}

We consider the test images with the lowest strength of additive Gaussian noise as in-distribution data since some Gaussian noise was also added during the image generation. All probabilistic deep learning methods achieve an accuracy of 1 except MC-dropout-0.4, which achieves an accuracy of $0.9988$.

Figure~\ref{fig:uncertainty_accuracy} provides an overview of the character accuracy of the probabilistic deep learning methods. We evaluate the four competing methods BatchEnsemble (blue), MC-dropout-0.1 (orange), MC-dropout-0.4 (green), and deep ensemble (red). The y-axis shows the character accuracy. The x-axis represents the strength of the degradation, with an increase in the degradation from left to right. Salt \& pepper (left) noise leads to lower accuracies than Gaussian noise (middle left). Although Fig.~\ref{fig:example_sp_01} shows that the license plate is still readable for probability $p = 0.1$, the accuracy drops already for $p > 0.01$. The deep ensemble performs best on images corrupted with Gaussian noise, while MC-dropout-0.4 achieves superior performance on salt \& pepper noise. Horizontal blur (middle right) is a challenge for the probabilistic deep learning methods. Even for a small kernel size $k=3$, the accuracy is well below 1. Vertical blur (right) is less challenging for a small kernel size, but the accuracy rapidly decreases with increasing kernel size. MC-dropout-0.4 achieves the highest accuracy across all kernel sizes for both horizontal and vertical blur. While MC-dropout-0.1 and the deep ensemble achieve similar results, BatchEnsemble performs worst. Except for Gaussian noise, BatchEnsemble is outperformed by the competing probabilistic deep learning methods across all levels of degradations.

Table~\ref{tab:uncertainty_accuracy} shows the character accuracy for each degradation type averaged over all degradation strengths. The results confirm the observations stated above. Salt \& pepper noise is more challenging than additive Gaussian noise, except for MC-dropout-0.4. Blur is even more challenging. MC-dropout-0.4 achieves 0.524 and 0.55 accuracy on horizontal and vertical blur, respectively.

{ Table~\ref{tab:mixed} shows the mean character accuracy for mixed types of degradations. We combine Gaussian noise with horizontal blur (left), Gaussian noise with defocus blur (middle), and salt \& pepper noise with defocus blur (right). In this experiment, we use a fixed degradation strength and compute the mean accuracy on the DS-Full dataset. MC-dropout-0.4 achieves the highest accuracy in all three cases. In the previous experiment, the ensemble performed best with Gaussian noise on all degradation levels. However, with little Gaussian noise and additional horizontal blur, the accuracy drops considerably. Defocus blur with Gaussian noise or salt \& pepper noise substantially lowers the accuracy of the models. Here, MC-Dropout-0.4 performs by far the best.
}

\begin{figure*}[!tbp]	
  \setlength \figurewidth{0.19\textwidth}
  \setlength \figureheight{1.0\figurewidth}
  \scriptsize
  
   \centering
  \includegraphics[width=0.6\textwidth]{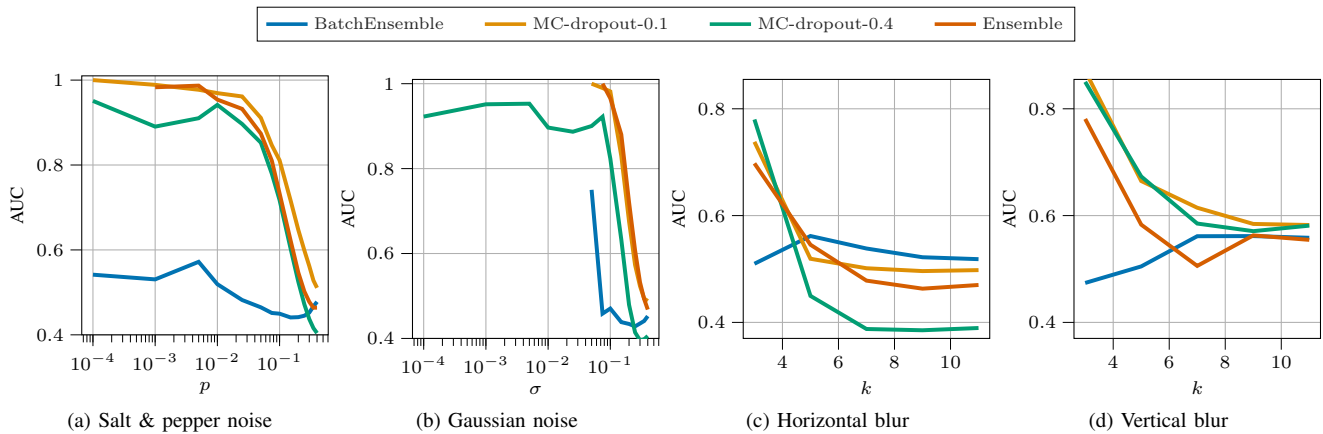}

  \subfloat[Salt \& pepper noise\label{fig:uncertainty_noise_auc_sp}]{\input{tex/noise-uncertainty-sp-AUC-pr-5}}  
  \subfloat[Gaussian noise \label{fig:uncertainty_noise_auc_gaussian}]{\input{tex/noise-uncertainty-gaussian-AUC-pr-5}}
 \subfloat[Horizontal blur \label{fig:uncertainty_blur_auc_horizontal}]{\input{tex/blur-uncertainty-horizontal-AUC-pr-5}}
 \subfloat[Vertical blur \label{fig:uncertainty_blur_auc_vertical}]{\input{tex/blur-uncertainty-vertical-AUC-pr-5}}

 \caption{AUC on the DS-Full dataset. The AUC of the precision-recall curve of the BatchEnsemble, MC-dropout, and the deep ensemble on the synthetic test dataset. The low-resolution images are corrupted with horizontal or vertical blur with varying kernel size $k$. Moreover, Gaussian noise with standard deviation $\sigma$ or salt \& pepper noise with probability $p$ is added to the low-resolution images.}
 \label{fig:uncertainty_auc}
\end{figure*}

Figure~\ref{fig:example_lr} shows that the license plate numbers in the training and validation images are still readable. In contrast, the strongly degraded test images in Fig.~\ref{fig:example_g1} - Fig.~\ref{fig:example_v3} are mostly undecipherable. Thus, the probabilistic deep learning methods are robust to unseen image degradations.

\subsubsection{Detection of false predictions}

However, on-par recognition performance is only one property of probabilistic deep learning methods. An even larger benefit may be gained from their intrinsic ability to detect false predictions. Figure~\ref{fig:uncertainty_auc} reports the AUCs for detecting false predictions from the magnitude of the uncertainty on the DS-Full dataset. We visualize the results for salt \& pepper noise (left), additive Gaussian noise (middle left), horizontal blur (middle right), and vertical blur (right)  for increasing strength of degradation. The y-axis shows the AUC. The x-axis visualizes the increasing strength of the degradation. The line plots start at those points where the model no longer reach an accuracy of 1.

Since salt \& pepper noise is more challenging, misclassifications already occur for $p > 10^{-4}$, hence we start to report the AUCs from this. The competing methods, except for BatchEnsemble, achieve a stable AUC around 0.96 and 0.92 for MC-dropout-0.1 and MC-dropout-0.4, respectively. When additive Gaussian noise is present, the competing methods expect MC-dropout-0.4 achieve an accuracy of 1 for $\sigma < 0.05$ or $\sigma < 0.075$. Thus, we can not compute an AUC. MC-dropout-0.4 has a stable AUC in the range of 0.9 to 0.95 for $0.0001 \leq \sigma \leq 0.75$. From this point, however, all AUC values sharply drop. For horizontal and vertical blur, the methods exhibit similar behavior. The AUC drops drastically for $k > 3$ by about 0.2 to 0.3. It can be observed that BatchEnsemble generally performs poorly and mostly achieves an AUC of around 0.5.

\begin{table}[!tbp]
	\centering
	\caption{AUC on the DS-Hard dataset. Mean AUC of the precision-recall curve for the BatchEnsemble, MC-dropout, and the deep ensemble.}
   \begin{tabular}{lrrrr} 
    &  \multicolumn{2}{c}{noise} &  \multicolumn{2}{c}{blur}  \\
    \cmidrule(lr){2-3}\cmidrule(lr){4-5} 
    Method &  salt \& pepper &  Gaussian & horizontal & vertical \\ \hline
    BatchEnsemble & 0.453 & 0.443 & 0.537 & 0.513 \\
    MC-dropout-0.1 &\textbf{0.625} & 0.652  &\textbf{0.586} &\textbf{0.717} \\
    MC-dropout-0.4 & 0.511 & 0.507 & 0.539 & 0.703 \\
    Ensemble & 0.544 &\textbf{0.667} & 0.574 & 0.623 \\
     \end{tabular} 
     \label{tab:uncertainty_auc_part}
    \end{table}

All competing methods struggle with strongly degraded images. Table~\ref{tab:uncertainty_auc_part} reports the mean AUC for these strongly degraded images on the DS-Hard dataset. For salt \& pepper noise (left), MC-dropout-0.1 best separates the predictive uncertainty values of false and correct predictions. The deep ensemble achieves the highest AUC on additive Gaussian noise (middle left) with $0.667$. For all other degradation types, MC-dropout-0.1 achieves the best results. Vertical blur (right) and additive Gaussian noise follow closely. The worst results are obtained when salt \& pepper noise ($0.625$) and horizontal blur ($0.586$) are present in the images. With this experiment, we aim at visualizing the border-line case. The test images with strong degradation vary greatly from the clean low-resolution images in the training data. In a real world scenario, the difference might not be that big. Therefore, the AUC values should not be seen absolutely but relatively as a comparison of the methods. But even in the border-line case, the competing methods except BatchEnsemble provide significantly better results than guessing.

To conclude the experiment, we assess the probabilistic deep learning techniques. MC-dropout-0.1 produces the most reliable detection of false predictions, closely followed by the deep ensemble. In FLPR, the informative value of the predictive uncertainty is important. Since no verification is possible, well-separated predictive uncertainties of correct and wrong predictions are important. Thus, we suggest MC-dropout with a lower dropout rate in this scenario. MC-dropout-0.4 and the deep ensemble achieve superior character recognition accuracy. For ALPR, high accuracy is important. Here, MC-dropout with a higher dropout rate or the deep ensemble is a better choice. BatchEnsemble is outperformed across all degradations and, therefore, not recommended.

 \subsection{Ablation on MC-dropout: }


%

 In this Section, we investigate influencing factors for the reliability of probabilistic deep learning techniques. In the previous experiment, the deep ensemble and MC-dropout performed best in terms of character and detection of false predictions. Due to the lower memory requirements and widespread usage, we only use MC-dropout for this experiment.

 Typically, increasing the dropout rate while keeping the number of trainable parameters fixed leads to the following trends. The accuracy decreases with an increasing dropout rate, while the predictive uncertainty of correct and wrong predictions becomes better separable with the increasing dropout rate. In contrast, we adapt the size of the neural network to the dropout rate such that the number of the parameter is fixed after dropout is applied.

First, we investigate the influence of the dropout rate and the number of inference runs on the accuracy and second on the uncertainty performance of MC-dropout with the baseline LPR CNN. Third, we examine changes in the distributions of the predictive uncertainty of correct and false predictions for an increasing number of inference runs and strengths of degradation. Narrow distributions that are well separable are ideal for the reliable detection of false predictions.

When the size of the neural network is adapted, we observe an inverted behavior of MC-dropout. The accuracy increases with the increasing dropout rate. The detection of misclassifications is better for lower dropout rates. The distributions of the predictive uncertainty of correct and false predictions converge with the increasing number of inference runs. The same behavior is observed for increasing the strength of degradation. Thus, a smaller difference between training and test distribution allows more reliable detection of misclassifications.

\begin{table}[!tbp]
  \centering
  \caption{Accuracy on the DS-Full dataset. Mean character recognition accuracy of MC-dropout with 5 and 50 inference runs and varying dropout rate. For each degradation type, the values represent the mean accuracy on all degradation levels.}
\begin{tabular}{lrrrrrr} 
  $r$  & & 0.1 & 0.2 & 0.3 & 0.4 & 0.5 \\ \hline
   \multirow{2}{*}{salt \& pepper} & 5 & 0.778 & 0.818 & 0.832 &\textbf{0.844} & 0.592 \\
          & 50 & 0.792  & 0.841 & 0.860 &\textbf{0.883} & 0.599 \\ 
    \multirow{2}{*}{Gaussian} & 5 & 0.804 &\textbf{0.814} & 0.802 & 0.802 & 0.696 \\
          & 50 & 0.815 &\textbf{0.835} & 0.821 & 0.825 & 0.700  \\
    \multirow{2}{*}{horizontal} & 5 & 0.292 & 0.352 & 0.449 &\textbf{0.517} & 0.177 \\
           &   50 & 0.299 & 0.355 & 0.461 &\textbf{0.550} & 0.178 \\
    \multirow{2}{*}{vertical} & 5 & 0.368 & 0.373 & 0.444 &\textbf{0.548} & 0.172 \\
     & 50  & 0.370 & 0.377  & 0.450 &\textbf{0.567} & 0.175 \\               
   \end{tabular}
   \label{tab:dropout_accuracy}
\end{table} 

\subsubsection{Character recognition} 
\label{sec:ablation_acc}
The models benefit from the increase in inference runs. Table~\ref{tab:dropout_accuracy} shows the mean character recognition accuracy for 5 and 50 inference runs and varying dropout rates on the DS-Full dataset. When salt \& pepper noise (left) corrupts the image, the performance of MC-dropout increases with increasing dropout rate until dropout rate $r = 0.4$. Applying more dropout in the CNN results in a strong performance decrease. Additionally, MC-dropout with higher dropout rates benefits more from the increasing number of inference runs. The accuracy of MC-dropout-0.4 improves by nearly 5\% with more inference runs. In contrast, the improvement is only 2\% for the model with dropout rate $r = 0.1$. Lower dropout rates are best suited for additive Gaussian noise (middle left). Horizontal (middle right) and vertical (right) blur behave similarly as salt \& pepper noise, but the performance gain with increasing dropout rate is larger. For example, with 50 inference runs on images corrupted by vertical blur, MC-dropout-0.4 achieves an accuracy of $0.567$. The model's accuracy with dropout rate $r = 0.3$ is $0.450$.

\begin{table}[!tbp]
	\centering
	\caption{AUC on the DS-Hard dataset. Mean precision-recall AUC  of MC-dropout with 5 and 50 inference runs and varying dropout rate.}
  \begin{tabular}{lrrrrrr}  
    $r$ &  & 0.1 & 0.2 & 0.3 & 0.4 & 0.5 \\ \hline
    \multirow{2}{*}{salt \& pepper} & 5 &\textbf{0.623} & 0.603 & 0.558 & 0.517 & 0.540 \\
    & 50 &\textbf{0.609} & 0.565 & 0.517 & 0.462 & 0.533 \\
    \multirow{2}{*}{Gaussian} & 5 &\textbf{0.648} & 0.604 & 0.559  & 0.511 & 0.597 \\
    & 50 &\textbf{0.624} & 0.580 & 0.527 & 0.455 & 0.589 \\
    \multirow{2}{*}{horizontal} & 5 & 0.581 &\textbf{0.615}  &\textbf{0.615}  & 0.553 & 0.498 \\
    & 50 & 0.581 &\textbf{0.608} & 0.585 & 0.502 & 0.464 \\
    \multirow{2}{*}{vertical} & 5 &\textbf{0.716} & 0.689 & 0.712 & 0.690 & 0.519 \\
    & 50 &\textbf{0.748} & 0.716 & 0.747  & 0.731 & 0.497 \\
   \end{tabular}
   \label{tab:dropout_auc_part}
\end{table}

\begin{figure*}[!tbp]
  \setlength \figurewidth{0.2\textwidth}
  \setlength \figureheight{0.8\figurewidth}
  \scriptsize
   \centering
  \includegraphics[width=0.7\textwidth]{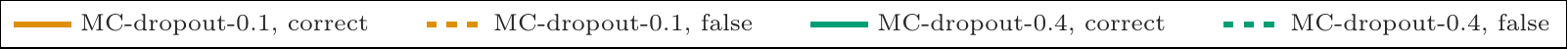}

  \subfloat[Salt \& pepper noise ($p = 0.01$)\label{fig:dropout_median_sp_001}]{\input{tex/noise-sp-median-001}}
  \subfloat[Gaussian noise ($\sigma = 0.2$)\label{fig:dropout_median_gaussian_02}]{\input{tex/noise-gaussian-median-02}}
 \subfloat[Horizontal blur ($k = 5$)\label{fig:dropout_median_horizontal_5}]{\input{tex/blur-horizontal-median-5}}

 \caption{Median of the predictive uncertainty values displayed for increasing number of inference runs. The dashed lines show the median predictive uncertainty values of falsely classified characters, and the solid lines the median predictive uncertainty values of correctly classified characters.}
 \label{fig:dropout_median}
\end{figure*}

\begin{figure*}[!tbp]
  \setlength \figurewidth{0.2\textwidth}
  \setlength \figureheight{0.8\figurewidth}
  \scriptsize
   \centering
  \includegraphics[width=0.7\textwidth]{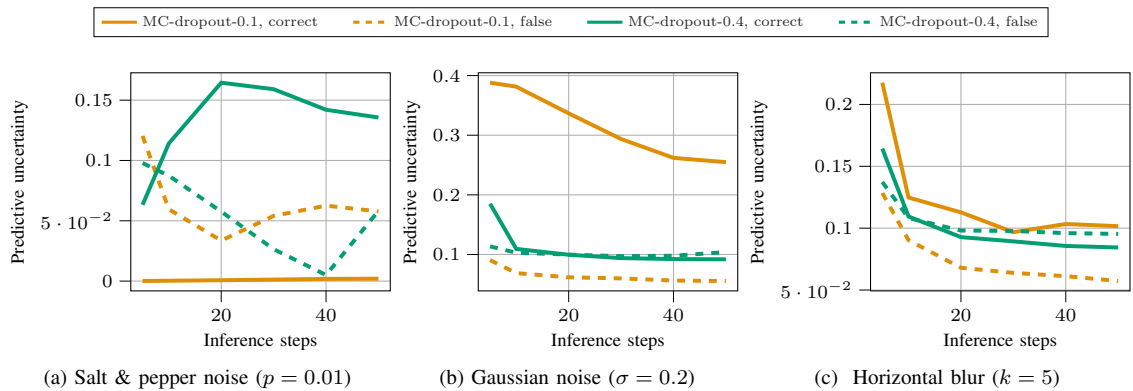}
  
  \subfloat[Salt \& pepper noise ($p = 0.01$) \label{fig:dropout_spread_sp_001}]{\input{tex/noise-sp-spread-001}}
  \subfloat[Gaussian noise ($\sigma = 0.2$) \label{fig:dropout_spread_gaussian_02}]{\input{tex/noise-gaussian-spread-02}}
  \subfloat[ Horizontal blur ($k = 5$) \label{fig:dropout_spread_horizontal_5}]{\input{tex/blur-horizontal-spread-5}}

  \caption{ Interquartile range of predictive uncertainty values displayed for increasing number of inference runs. The dashed lines show the interquartile range of the predictive uncertainty values of falsely classified characters and the solid lines the interquartile range of the predictive uncertainty values of correctly classified characters.}
  \label{fig:dropout_spread}
  \end{figure*}

\subsubsection{Detection of false predictions}
\label{sec:ablation_auc}

Both, ALPR and FLPR, require a reliable detection of false predictions. Table~\ref{tab:dropout_auc_part} visualizes the AUC on the DS-Hard dataset for 5 and 50 inference runs. False predictions raised by salt \& pepper noise (left) and additive Gaussian noise (middle left) are best detected with MC-dropout-0.1. 
Additionally, the AUC does not benefit from an increasing number of inference runs. The behavior of MC-dropout on images corrupted with horizontal blur (middle right) is slightly different. Here, MC-dropout-0.2 achieves the highest AUC. Vertical blur (right) provokes a similar behavior of MC-dropout as noise, but the AUC slightly increases with increasing dropout rate for $r \leq 0.4$. 

We conclude from Subsec.~\ref{sec:ablation_acc} and Subsec.~\ref{sec:ablation_auc} that ALPR and FLPR require two different tuning strategies. In FLPR, lower numbers of inference runs are advised. The criminal investigator is not able to visually verify the prediction of the neural network. Thus, reliable detection of false predictions is a desired feature of the license plate recognition CNN. In ALPR, the number of inference runs can be set to a larger value. Here, the accuracy of the license plate recognition is important.

\subsubsection{Evaluation of predictive uncertainty}





Two different factors cause a poor detection of false predictions. First, with an increasing number of inference runs, we observed a decrease in the AUC. Second, with the increasing strength of the degradation, the AUC decreases. This behavior indicates that the predictive uncertainties of correct and wrong predictions are poorly separable. Therefore, we examine the changes in the distribution of the predictive uncertainty values of correct (solid) and false (dashed) predictions for different inference runs and increasing strengths of degradation more closely. We use the median and the interquartile range to quantize the distribution. The predictive uncertainties are computed on a subset of \numprint{1000} test images. The experiments are conducted with MC-dropout-0.1 (orange) and MC-dropout-0.4 (green), since these models achieve the highest AUC and accuracy, respectively.

To visualize the influence of inference runs on the detection of false prediction, we choose three degradations with varying strength. While salt \& pepper noise with $p=0.01$ is less severe, additive Gaussian noise with $\sigma = 0.2$ and horizontal blur with $k = 5$ significantly lower the image quality. 

Figure~\ref{fig:dropout_median} visualizes the median predictive uncertainty of correct (M$_\textnormal{c}$) and false (M$_\textnormal{f}$) predictions for different inference runs. The y-axis shows median predictive uncertainty. The x-axis shows the number of inference runs. For salt \& pepper noise (left) M$_\textnormal{c}$ and M$_\textnormal{f}$ increase with the number of inference runs. However, M$_\textnormal{f}$ is more stable than M$_\textnormal{c}$. The median predictive uncertainties are nicely separated. To identify false predictions better than just guessing, the median predictive uncertainty of false predictions has to be above that of correct predictions. This does not always hold for severely degraded images with additive Gaussian noise (middle) and horizontal blur (right). Here, MC-dropout-0.1 allows better separation than MC-dropout-0.4.

We assume the increase of the median predictive uncertainty with an increasing number of inference runs is linked to the increased accuracy that is observed in Tab.~\ref{tab:dropout_accuracy}. Some characters predicted wrongly with five inference runs become correct with 50 inference runs. However, the predictive uncertainty for those characters is still high, which increases the overall mean uncertainty.

In addition to widely spaced distributions, narrow distributions are also important to detect misclassifications reliably. Figure~\ref{fig:dropout_spread} visualizes the interquartile range of the predictive uncertainty values of correct (solid) and false (dashed) predictions. The y-axis shows the interquartile range of the predictive uncertainty value. The x-axis shows the number of inference runs. When little salt \& pepper noise (left) is present, the interquartile range is smaller compared to severely degraded images with Gaussian noise (middle) and horizontal blur (right). In general, the interquartile range tends to decrease with an increasing number of inference runs. An exception to this behavior is the spread of the predictive uncertainty values of correct predictions from MC-dropout-0.4, which increases.

\begin{figure*}[!tbp]
  \setlength \figurewidth{0.19\textwidth}
  \setlength \figureheight{0.8\figurewidth}
  \scriptsize
   \centering
  \includegraphics[width=0.7\textwidth]{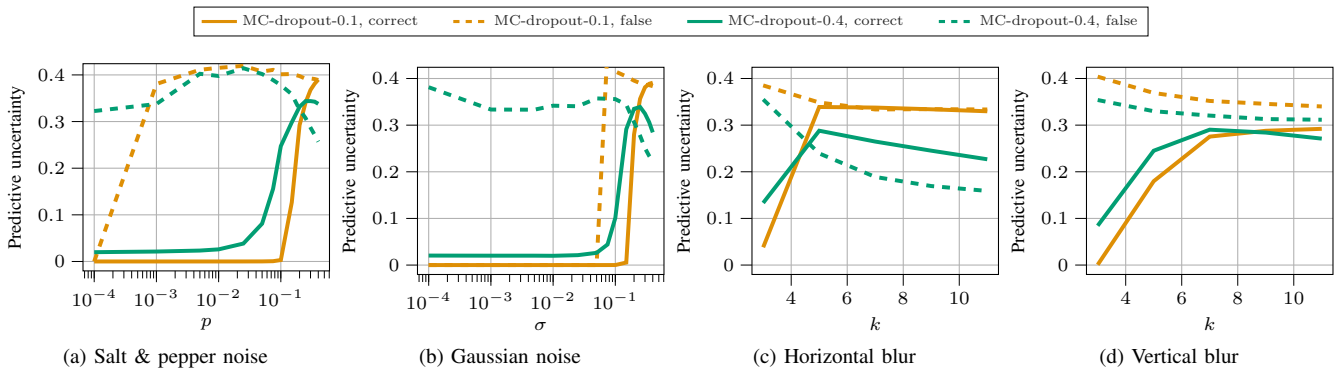}

 \subfloat[Salt \& pepper noise \label{fig:dropout_median_sp}]{\input{tex/noise-sp-median}}
 \subfloat[Gaussian noise \label{fig:dropout_median_gaussian}]{\input{tex/noise-gaussian-median}}
 \subfloat[Horizontal blur \label{fig:dropout_median_vertical}]{\input{tex/blur-horizontal-median}}
 \subfloat[Vertical blur \label{fig:dropout_median_horizontal}]{\input{tex/blur-vertical-median}}

 \caption{Median of the predictive uncertainty values of MC-dropout displayed for different degradation levels with five inference runs. The dashed lines show the median predictive uncertainty values of falsely classified characters, and the solid lines the median predictive uncertainty values of correctly classified characters.}
 \label{fig:dropout_median_threshold}
\end{figure*}

Besides the number of inference runs, the strength of the degradation influences the detection of false predictions. Figure~\ref{fig:dropout_median_threshold} visualize the median predictive uncertainty of correct (solid) and false (dashed) predictions of MC-dropout with ten inference runs. The y-axis shows median predictive uncertainty. The x-axis represents the strength of the degradation, with an increase in the degradation from left to right. We choose ten runs as a tradeoff between decreasing spread and increasing median predictive uncertainty of correct predictions. For salt \& pepper noise (left) and additive Gaussian noise (middle left), MC-dropout-0.4 has higher predictive uncertainty values of falsely classified characters on higher quality images. However, with increasing degradation strength, the median predictive uncertainty of wrong characters becomes lower than that of correct characters. MC-dropout-0.1 is better suited for strongly degraded images but later indicates a wrong character. Except for horizontal blur (middle right), falsely classified characters' median predictive uncertainty values are above 0.3. This value can be a potential threshold in the context of ALPR. However, this threshold is not suitable for strongly degraded images.

 \subsection{Influence of Super-Resolution on Classification}

The generalization of neural networks to unseen degradations can also be addressed with multi-task learning~\cite{Caruana1993, Caruana1997, Ruder2017}: training multiple tasks on the same input image serves as an inductive bias. In this experiment, we investigate the inductive bias introduced by the super-resolution in combination with license plate recognition. We implement both tasks in the SR$^2$ framework~\cite{Schirrmacher2020}. In contrast to previous work~\cite{Hilario2017}, we arrange both tasks in parallel. The parallel arrangement minimizes error propagation and provides better character recognition when faced with out-of-distribution data, as shown by~\cite{Schirrmacher2020}.

The experiments prove the hypothesis that super-resolution introduces an inductive bias that is beneficial for character recognition. All probabilistic deep learning methods benefit from super-resolution in terms of both accuracy and predictive uncertainty. The performance boost is best seen on blurred images. 

\begin{table}[!tbp]
	\centering
	\caption{Accuracy on the DS-Full dataset. Mean character recognition accuracy of the BatchEnsemble, MC-dropout, and the deep ensemble. For each degradation type, the values represent the mean accuracy of all degradation levels. The table compares the LPR CNN to the SR$^2$ framework with $w_{lpr} = 20$ and $w_{sr} = 1$.  }
\begin{tabular}{lrrrr} 
  Method &  s \& p &  Gaussian & horizontal & vertical \\ \hline
  BatchEnsemble CNN & 0.691 & 0.832 & 0.230 & 0.280 \\
  BatchEnsemble SR$^2$& 0.720 & 0.852 & 0.482 & 0.393 \\
  MC-dropout-0.1 CNN & 0.775 & 0.803 & 0.295 & 0.371 \\
  MC-dropout-0.1 SR$^2$ & 0.748 & 0.842  & 0.549 & 0.457 \\
  MC-dropout-0.4 CNN  &\textbf{0.844} & 0.801 & 0.524 & \textbf{0.550}\\
  MC-dropout-0.4 SR$^2$ &\textbf{0.844} & 0.831 &\textbf{0.685} &0.531 \\
  Ensemble CNN & 0.726 & 0.846  & 0.306 & 0.357 \\
  Ensemble SR$^2$ & 0.791 &\textbf{0.910}& 0.538 & 0.407 \\
   \end{tabular}
   \label{tab:sr2_accuracy}
\end{table}

\subsubsection{Character Recognition}

Table~\ref{tab:sr2_accuracy} reports the mean character recognition accuracy on the DS-Full dataset. Super-resolution increases the character recognition accuracy on noisy images when salt \& pepper noise (left) is present, except for MC-dropout-0.1. MC-dropout-0.4 with both backbones performs best. Additive Gaussian noise (middle left) poses less of a problem for character recognition. The deep ensemble with SR$^2$ as backbone achieves the highest accuracy with $0.91$ and $0.99$, respectively. All probabilistic deep learning techniques benefit from the inductive bias of the super-resolution task. A significant increase in performance is observed for blurred images. The ensemble, for example, improves by 75\% on horizontally (middle right) blurred images. BatchEnsemble undergoes an even larger improvement of 109\%. When vertical blur (right) is present, MC-dropout-0.4 with the CNN as a backbone performs best. Thus, MC-dropout-0.4 does not benefit from the inductive bias introduced by super-resolution. The other probabilistic deep learning techniques benefit from super-resolution.

\begin{table}[!tbp]
	\centering
	\caption{AUC on the DS-Hard dataset. Mean AUC of the precision-recall curve for the BatchEnsemble, MC-dropout, and the deep ensemble.}
   \begin{tabular}{lrrrr} 
    Method &  s \& p &  Gaussian & horizontal & vertical \\ \hline
    BatchEnsemble CNN & 0.453 & 0.443  & 0.537 & 0.513 \\
    BatchEnsemble SR$^2$& 0.488 & 0.481  & 0.477 & 0.503 \\
    MC-dropout-0.1 CNN  & 0.625 & 0.652  & 0.586 & 0.717 \\
    MC-dropout-0.1 SR$^2$&\textbf{0.644} & 0.708  &\textbf{0.757} &\textbf{0.775} \\
    MC-dropout-0.4 CNN  & 0.511 & 0.507  & 0.539 & 0.703 \\
    MC-dropout-0.4 SR$^2$& 0.495 & 0.507  & 0.599 & 0.736 \\
    Ensemble CNN  & 0.544 & 0.667 & 0.574 & 0.623 \\
    Ensemble SR$^2$& 0.605 &\textbf{0.751} & 0.648 & 0.714 \\
     \end{tabular} 
     \label{tab:sr2_auc_part}
\end{table}

\subsubsection{Detection of false predictions}

Table~\ref{tab:sr2_auc_part} visualizes the mean AUC on the DS-Hard dataset. On images corrupted with salt \& pepper noise (left), only MC-dropout-0.4 does not benefit from the super-resolution. MC-dropout-0.1 with SR$^2$ as a backbone achieves the highest AUC. For additive Gaussian noise (middle left), the deep ensemble with SR$^2$ as a backbone achieves the highest AUC. The character recognition performance of all probabilistic deep learning techniques increases with SR$^2$ as a backbone. Horizontal (middle right) and vertical blur (right) are less challenging for the SR$^2$ backbone, except for BatchEnsemble. Again, MC-dropout-0.1 with SR$^2$ as a backbone achieves the highest AUC.

In conclusion, super-resolution improves the character recognition accuracy and detection of false predictions. In addition to the improvement, we see great potential for super-resolution to be used as an additional verification tool on images with a reduced quality. 

 \subsection{Comparison to Related Work}

The unique feature of probabilistic deep learning techniques is the quantization of predictive uncertainty. The previous experiments show the beneficial use of predictive uncertainty for detecting misclassifications, which has not been explored yet for license plate recognition. Predictive uncertainty can indicate where the prediction is not reliable and should be verified. Still, we make a ranking of the character recognition performance compared to related work. This section compares the vanilla LPR CNN and MC-dropout with LPR CNN as a backbone to the robust attentional framework proposed by~\cite{Zhang2020}, MANGO~\cite{Qiao2020}, and RPnet~\cite{Xu2018}. 

 \begin{table}[!tbp]
	\centering
	\caption{AUC on the CCPD dataset. Mean character and license plate recognition accuracy on the CCPD base test dataset. For MC-dropout, the AUC of the precision-recall curve is provided. The number of inference runs is set to 50 for MC-dropout.}
\begin{tabular}{L{2.7cm}R{1.5cm}R{1.5cm}r}
Method & character accuracy&  license plate accuracy & AUC \\ \hline
RPnet~\cite{Xu2018}  &-&0.993&- \\
MANGO~\cite{Qiao2020}  &-&0.990&- \\
Zhang \etal~\cite{Zhang2020} (real)  &-& 0.996&- \\
Zhang \etal~\cite{Zhang2020} (real \& synthetic)  &-& \textbf{0.998}&- \\
LPR CNN &0.998&0.989&-\\
MC-dropout-0.1 & 0.999 & 0.992& 0.993 \\
MC-dropout-0.2 & 0.999 & 0.992& 0.993 \\
MC-dropout-0.3 & 0.998 & 0.988& 0.995 \\
MC-dropout-0.4 & 0.998 & 0.990& 0.994 \\
 \end{tabular}
 \label{tab:realWorld}

\end{table}
 
 With this comparison, we aim at ranking the LPR CNN on the real-world CCPD dataset. Since the focus of this paper is on license plate recognition, we omit the license plate detection step, following the findings by Zhang \etal~\cite{Zhang2020}. In an ablation study, the authors compared the recognition accuracy using ground truth bounding boxes to the recognition accuracy with the license plate detection performed with YOLO. On the base test set, there is no difference in the accuracies. Thus, we can compare the license plate recognition accuracy of the LPR CNN with ground truth bounding boxes to the accuracies reported in the related work. We show the results reported in~\cite{Xu2018}, since~\cite{Zhang2020} and ~\cite{Qiao2020} report slightly different accuracies of RPnet on the base dataset than the original paper.

 Table~\ref{tab:realWorld} shows the results for LPR CNN and MC-dropout along with the results of the competing methods. We report the accuracy of predicting a character (left) correctly and predicting the whole license plate (middle) correctly. The character accuracy is only evaluated for the LPR CNN and MC-dropout. The accuracy of the methods is close to 1. The robust attentional framework achieves the highest license plate accuracy when trained on real \& synthetic data. While Xu \etal~\cite{Xu2018} perform slightly better than the proposed LPR CNN, MC-dropout-0.1 and MC-dropout-0.2 outperform MANGO~\cite{Qiao2020}.

In ALPR, there is no manual verification of the predictions. All competing methods make false predictions, but none of them can detect of these misclassifications. In contrast, MC-dropout allows quantifying the uncertainty of the prediction. MC-dropout-0.3 achieves an AUC of $0.995$. Therefore, we argue that this property compensates for the slightly lower license plate recognition accuracy.

\begin{figure}[!tbp]

  \begin{tikzpicture}
    
    \node at (0,0) {\pgfimage[width = 0.35\textwidth]{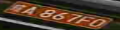}};
    \node at (4.25,0.65)  {\begin{CJK*}{UTF8}{gbsn}皖\end{CJK*}A88X00};
    \node at (4.25,0.35)  {\begin{CJK*}{UTF8}{gbsn}皖\end{CJK*}A86Z90};
    \node at (4.25,0.0)   {\begin{CJK*}{UTF8}{gbsn}皖\end{CJK*}A86758};
    \node at (4.25,-0.35) {\begin{CJK*}{UTF8}{gbsn}皖\end{CJK*}A86700};
    \node at (4.25,-0.65) {\begin{CJK*}{UTF8}{gbsn}皖\end{CJK*}A86708};

    \node at (0,-2) {\pgfimage[width = 0.35\textwidth]{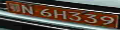}};
    \node at (4.25,-1.65) {\begin{CJK*}{UTF8}{gbsn}鄂\end{CJK*}N6H339};
    \node at (4.25,-1.35) {\begin{CJK*}{UTF8}{gbsn}豫\end{CJK*}N6H339};
    \node at (4.25,-2.0)  {\begin{CJK*}{UTF8}{gbsn}鄂\end{CJK*}N6H339};
    \node at (4.25,-2.35) {\begin{CJK*}{UTF8}{gbsn}鄂\end{CJK*}N6H339};
    \node at (4.25,-2.65) {\begin{CJK*}{UTF8}{gbsn}豫\end{CJK*}N6H339};

  \end{tikzpicture}

  \caption{Challenging example images from the CCPD base dataset. We show the input images (left) and the predictions of five inference runs of MC-dropout-0.1 (right). Here, the predictive uncertainty is beneficial to detect misclassifications.}
  \label{fig:CCPDExamples}
\end{figure}
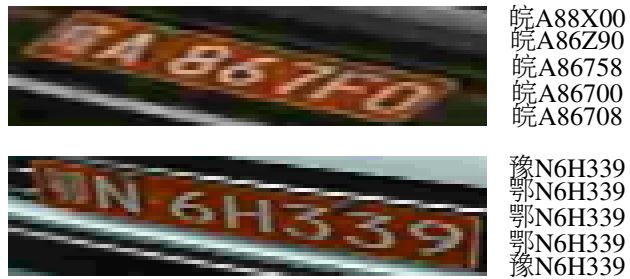

{ Figure~\ref{fig:CCPDExamples} highlights the advantages of MC-dropout-0.1 when faced with two challenging input images (left). We provide the predictions of five inference runs for each input (right). In these cases, the strong rotation (top) and the low resolution (bottom) impede the recognition performance of the network. We can identify the challenging parts of the license plate due to the difference in the predicted characters between inference runs.}

\section{Conclusion}
\label{sec:conclusion}

This paper proposes to model uncertainty for the task of license plate recognition explicitly. To the best of our knowledge, this has not been explored yet but offers helpful features for license plate recognition. For example, we demonstrate that the quantification of the prediction uncertainty allows the detection of misclassifications. We identify automatic license plate recognition and forensic license plate recognition as applications that benefit from predictive uncertainty.

We investigate three well-known probabilistic deep learning methods that quantify predictive uncertainty: BatchEnsemble, MC-dropout, and deep ensemble. Two neural network architectures are the backbones for these techniques. A state-of-the-art license plate recognition CNN serves as a baseline backbone. To exploit the benefits of multi-task learning, we combine super-resolution and license plate recognition in the SR$^2$ framework as a second backbone.

License plate recognition in the wild is complex since images stem from various acquisition settings. One must always consider a lower quality of the test data than that of the training data. We propose probabilistic deep learning as a tool to detect when the data and thus the character recognition are less reliable. For this purpose, the models are trained on high-quality images and tested on noisy or blurred lower-quality images. Except for BatchEnsemble, all probabilistic deep learning methods provide reasonable uncertainty estimates even for severely degraded images. Even better results are obtained when license plate recognition is combined with super-resolution in the SR$^2$ framework. The SR$^2$ framework improves both character recognition accuracy and detection of false predictions. For the future, we see super-resolution as a tool to additionally verify the prediction of images with a reduced quality. Here, the predictive uncertainty obtained per pixel can help identify less reliable character predictions of the LPR CNN. The hyperparameter of MC-dropout allows setting a stronger focus on either character recognition performance or reliable detection of false predictions.

\bibliographystyle{IEEEtran}
\bibliography{IEEEabrv,references}



\begin{IEEEbiography}[{\includegraphics[width=1in,height=1.25in,clip,keepaspectratio]{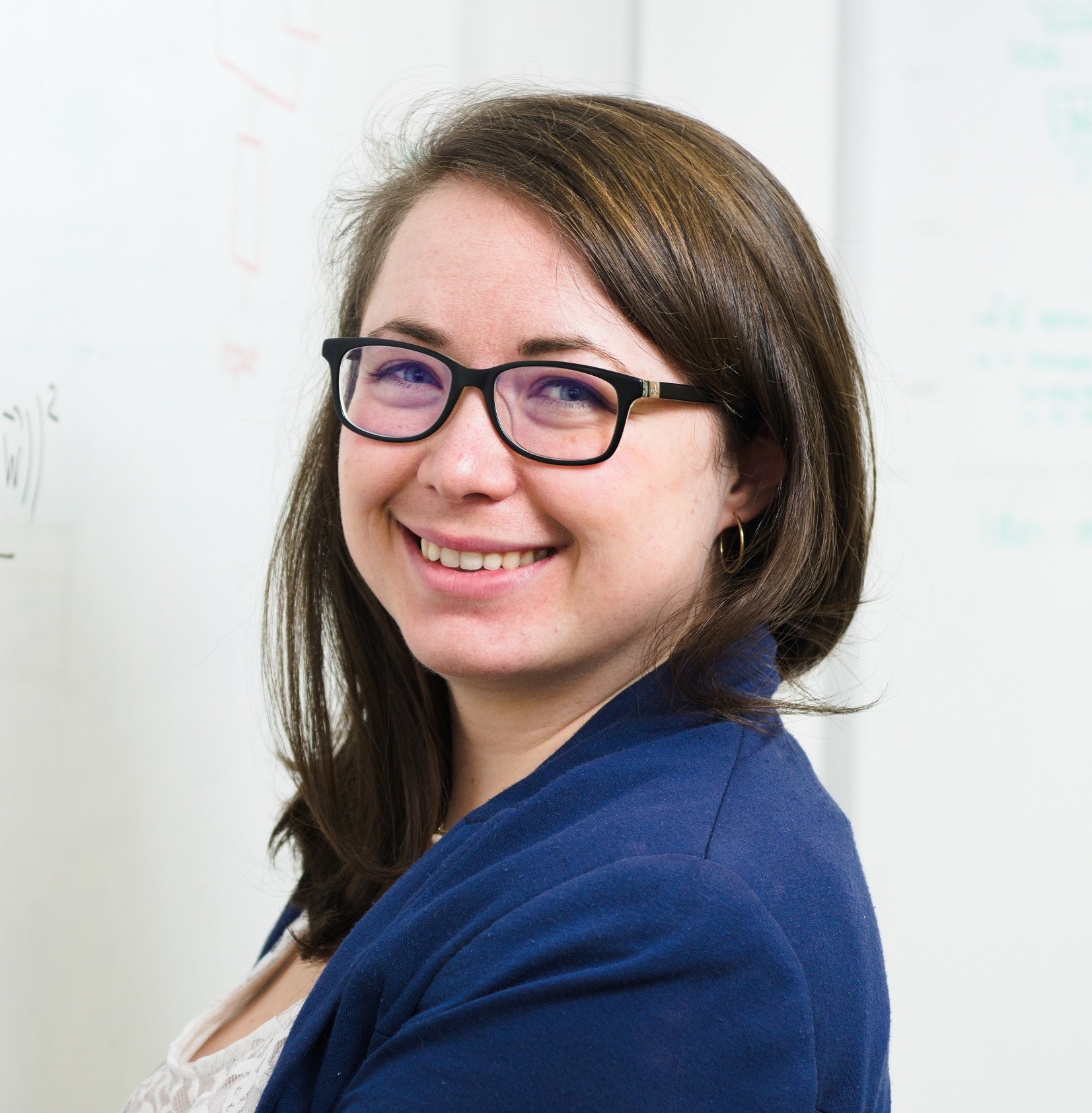}}]{Franziska Schirrmacher}
	received the M.Sc.\ degree in medical engineering from the Friedrich-Alexander University Erlangen-N\"urnberg (FAU), Erlangen, Germany, in 2017. From  2017  to  2019, she  was  a  researcher at  the  Pattern  Recognition  Lab  at  FAU. In 2019,  she  joined  the  IT Infrastructures Lab at FAU and is part of the Multimedia Security Group. Her research interests include image processing, machine learning, and image forensics.
\end{IEEEbiography}

\begin{IEEEbiography}[{\includegraphics[width=1in,height=1.25in,clip,keepaspectratio]{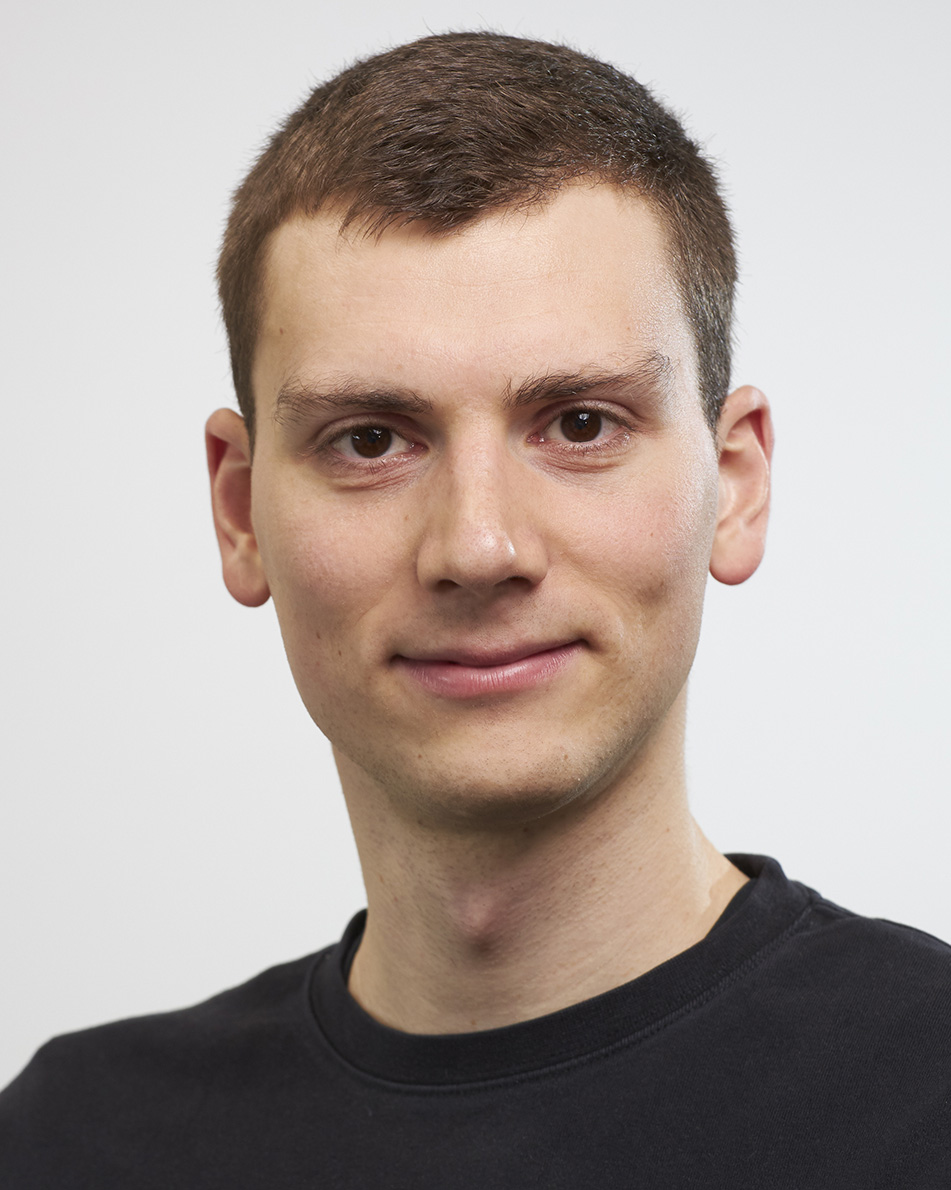}}]{Benedikt Lorch}
	received the M.Sc.\ degree in computer science from the Friedrich-Alexander University Erlangen-N\"urnberg (FAU), Erlangen, Germany, in 2018. In Septemer 2018 he joined the IT Infrastructures Security Lab as a Ph.D.\ student. His research interests include image forensics, computer vision, and machine learning.
\end{IEEEbiography}


\begin{IEEEbiography}[{\includegraphics[width=1in,height=1.25in,clip,keepaspectratio]{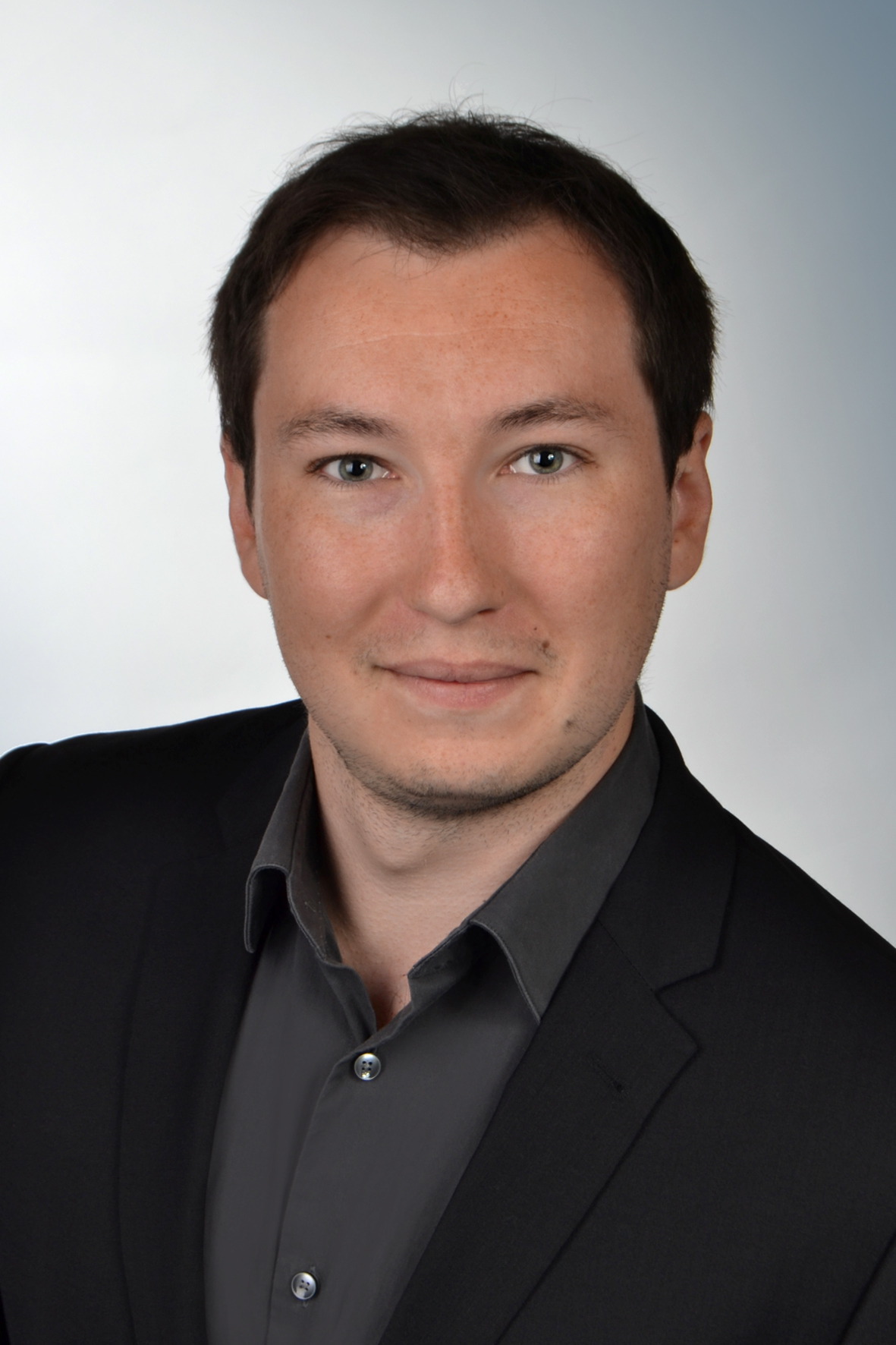}}]{Anatol Maier}
	received the M.Sc.\ degree in computer science from the Friedrich-Alexander University Erlangen-N\"urnberg (FAU), Erlangen, Germany, in 2019. Since November 2019, he is Ph.D. student at the IT Security Infrastructures Lab at FAU and part of the Multimedia Security Group. His research interests include reliable machine learning, deep probabilistic models, and computer vision with particular application in image and video forensics.
\end{IEEEbiography}

\begin{IEEEbiography}[{\includegraphics[width=1in,height=1.25in,clip,keepaspectratio]{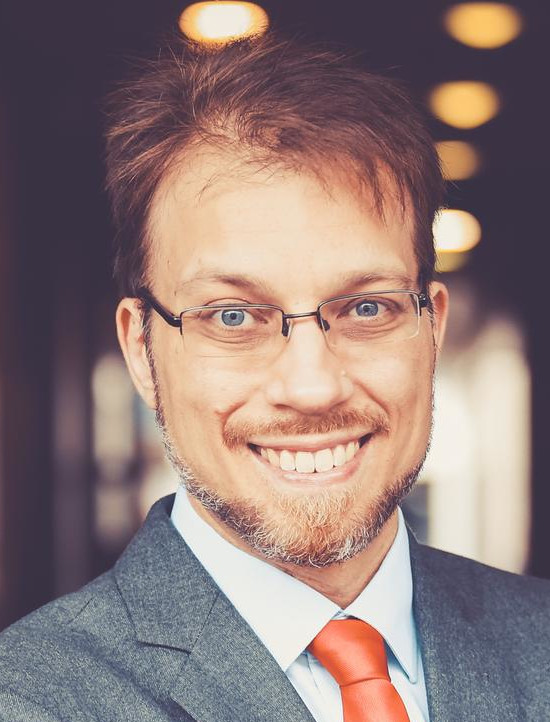}}]{Christian Riess}
  received the Ph.D. degree in computer science from the Friedrich-Alexander University Erlangen-N\"urnberg (FAU), Erlangen, Germany, in 2012. From 2013 to 2015, he was a Postdoc at the Radiological Sciences Laboratory, Stanford University, Stanford, CA, USA. Since 2015, he is the head of the Phase-Contrast X-ray Group at the Pattern Recognition Laboratory at FAU. Since 2016, he is senior researcher and head of the Multimedia Security Group at the IT Infrastructures Lab at FAU. He is currently a member of the IEEE Information Forensics and Security Technical Committee. His research interests include all aspects of image processing and imaging, particularly with applications in image and video forensics, X-ray phase contrast, color image processing, and computer vision.
  \end{IEEEbiography}

\end{document}

%% file: tex/noise-uncertainty-sp-aC-5.tex
\begin{tikzpicture}

\definecolor{color0}{rgb}{0.00392156862745098, 0.45098039215686275, 0.6980392156862745}
\definecolor{color1}{rgb}{0.8705882352941177, 0.5607843137254902, 0.019607843137254}
\definecolor{color2}{rgb}{0.00784313725490196, 0.6196078431372549, 0.45098039215686275}
\definecolor{color3}{rgb}{0.8352941176470589, 0.3686274509803922, 0.0}

\begin{axis}[
legend cell align={left},
legend style={
	fill opacity=0.8,
	draw opacity=1,
	text opacity=1,
	at={(0.5,1.22)},
	anchor=north,
	draw=white!80!black,
	legend columns=2
},
width=0.951\figurewidth,
height=\figureheight,
at={(0\figurewidth,0\figureheight)},
scale only axis,
ylabel near ticks,
xlabel near ticks,
log basis x={10},
tick align=outside,
tick pos=left,
xlabel={$p$},
ylabel={Character accuracy},
x grid style={white!69.0196078431373!black},
xmin=6.60536773049804e-03, xmax=0.605568102065137,
xmode=log,
xtick style={color=black},
y grid style={white!69.0196078431373!black},
ymin=0, ymax=1.01,
ytick style={color=black},
xmajorgrids,
ymajorgrids,
every axis plot/.append style={thick}
]
\addplot [semithick, color0,  line width=1.5pt]
table {%
0.01 0.992528571428572
0.025 0.9749
0.05 0.930557142857143
0.075 0.868385714285714
0.1 0.792728571428571
0.15 0.627357142857143
0.2 0.478814285714286
0.25 0.356557142857143
0.3 0.270814285714286
0.35 0.209128571428571
0.4 0.169328571428571
};
\addplot [semithick, color1,  line width=1.5pt]
table {%
0.01 0.9947
0.025 0.982942857142857
0.05 0.958671428571429
0.075 0.923657142857143
0.1 0.879857142857143
0.15 0.776457142857143
0.2 0.661914285714286
0.25 0.550728571428571
0.3 0.452314285714286
0.35 0.3694
0.4 0.300214285714286
};
\addplot [semithick, color2,  line width=1.5pt]
table {%
0.01 0.995828571428571
0.025 0.989371428571429
0.05 0.976514285714286
0.075 0.959557142857143
0.1 0.938085714285714
0.15 0.878657142857143
0.2 0.799585714285714
0.25 0.710371428571429
0.3 0.612428571428571
0.35 0.520971428571429
0.4 0.433914285714286
};
\addplot [semithick, color3,  line width=1.5pt]
table {%
0.01 0.995214285714286
0.025 0.984114285714286
0.05 0.954028571428571
0.075 0.909157142857143
0.1 0.850342857142857
0.15 0.708028571428571
0.2 0.560957142857143
0.25 0.431628571428571
0.3 0.326142857142857
0.35 0.250014285714286
0.4 0.195885714285714
};
\end{axis}

\end{tikzpicture}

%% file: tex/noise-uncertainty-gaussian-aC-5.tex
\begin{tikzpicture}

\definecolor{color0}{rgb}{0.00392156862745098, 0.45098039215686275, 0.6980392156862745}
\definecolor{color1}{rgb}{0.8705882352941177, 0.5607843137254902, 0.019607843137254}
\definecolor{color2}{rgb}{0.00784313725490196, 0.6196078431372549, 0.45098039215686275}
\definecolor{color3}{rgb}{0.8352941176470589, 0.3686274509803922, 0.0}

\begin{axis}[
legend cell align={left},
legend style={
	fill opacity=0.8,
	draw opacity=1,
	text opacity=1,
	at={(0.5,1.22)},
	anchor=north,
	draw=white!80!black,
	legend columns=2
},
width=0.951\figurewidth,
height=\figureheight,
at={(0\figurewidth,0\figureheight)},
scale only axis,
ylabel near ticks,
xlabel near ticks,
log basis x={10},
tick align=outside,
xlabel={$\sigma$},
ylabel={Character accuracy},
tick pos=left,
x grid style={white!69.0196078431373!black},
xmin=4.60536773049804e-02, xmax=0.605568102065137,
xmode=log,
xtick style={color=black},
y grid style={white!69.0196078431373!black},
ymin=0.0, ymax=1.01,
ytick style={color=black},
xmajorgrids,
ymajorgrids,
every axis plot/.append style={thick}
]
\addplot [semithick, color0,  line width=1.5pt]
table {%
0.05 0.999914285714286
0.075 0.999014285714286
0.1 0.993614285714286
0.15 0.948071428571429
0.2 0.836857142857143
0.25 0.676585714285714
0.3 0.5149
0.35 0.3868
0.4 0.293757142857143
};
\addplot [semithick, color1,  line width=1.5pt]
table {%
0.05 0.999928571428572
0.075 0.999485714285714
0.1 0.995585714285714
0.15 0.942685714285714
0.2 0.7828
0.25 0.578057142857143
0.3 0.416971428571429
0.35 0.301157142857143
0.4 0.2305
};
\addplot [semithick, color2,  line width=1.5pt]
table {%
0.05 0.998342857142857
0.075 0.9961
0.1 0.988842857142857
0.15 0.931042857142857
0.2 0.783414285714286
0.25 0.588928571428571
0.3 0.417228571428571
0.35 0.296914285714286
0.4 0.222528571428571
};
\addplot [semithick, color3,  line width=1.5pt]
table {%
0.05 1
0.075 0.999585714285714
0.1 0.996514285714286
0.15 0.964985714285714
0.2 0.869
0.25 0.719371428571429
0.3 0.5577
0.35 0.418114285714286
0.4 0.318642857142857
};
\end{axis}

\end{tikzpicture}

%% file: tex/blur-uncertainty-horizontal-aC-5.tex
\begin{tikzpicture}
\definecolor{color0}{rgb}{0.00392156862745098, 0.45098039215686275, 0.6980392156862745}
\definecolor{color1}{rgb}{0.8705882352941177, 0.5607843137254902, 0.019607843137254}
\definecolor{color2}{rgb}{0.00784313725490196, 0.6196078431372549, 0.45098039215686275}
\definecolor{color3}{rgb}{0.8352941176470589, 0.3686274509803922, 0.0}

\begin{axis}[
legend cell align={left},
legend style={
	fill opacity=0.8,
	draw opacity=1,
	text opacity=1,
	at={(0.5,1.22)},
	anchor=north,
	draw=white!80!black,
	legend columns=2
},
width=0.951\figurewidth,
height=\figureheight,
at={(0\figurewidth,0\figureheight)},
scale only axis,
ylabel near ticks,
xlabel near ticks,
tick align=outside,
tick pos=left,
xlabel={$k$},
ylabel={Character accuracy},
x grid style={white!69.0196078431373!black},
xmin=2.6, xmax=11.4,
xtick style={color=black},
y grid style={white!69.0196078431373!black},
ymin=0, ymax=1.01,
ytick style={color=black},
xmajorgrids,
ymajorgrids,
every axis plot/.append style={thick}
]
\addplot [semithick, color0,  line width=1.5pt]
table {%
3 0.581157142857143
5 0.182585714285714
7 0.138414285714286
9 0.126514285714286
11 0.1197
};
\addplot [semithick, color1,  line width=1.5pt]
table {%
3 0.814157142857143
5 0.230314285714286
7 0.157014285714286
9 0.143914285714286
11 0.130642857142857
};
\addplot [semithick, color2,  line width=1.5pt]
table {%
3 0.9227
5 0.6565
7 0.447357142857143
9 0.334814285714286
11 0.256771428571428
};
\addplot [semithick, color3,  line width=1.5pt]
table {%
3 0.658571428571428
5 0.308342857142857
7 0.214428571428571
9 0.190757142857143
11 0.158042857142857
};
\end{axis}

\end{tikzpicture}

%% file: tex/blur-uncertainty-vertical-aC-5.tex
\begin{tikzpicture}

\definecolor{color0}{rgb}{0.00392156862745098, 0.45098039215686275, 0.6980392156862745}
\definecolor{color1}{rgb}{0.8705882352941177, 0.5607843137254902, 0.019607843137254}
\definecolor{color2}{rgb}{0.00784313725490196, 0.6196078431372549, 0.45098039215686275}
\definecolor{color3}{rgb}{0.8352941176470589, 0.3686274509803922, 0.0}

\begin{axis}[
legend cell align={left},
legend style={
	fill opacity=0.8,
	draw opacity=1,
	text opacity=1,
	at={(0.5,1.22)},
	anchor=north,
	draw=white!80!black,
	legend columns=2
},
width=0.951\figurewidth,
height=\figureheight,
at={(0\figurewidth,0\figureheight)},
scale only axis,
ylabel near ticks,
xlabel near ticks,
tick align=outside,
tick pos=left,
xlabel={$k$},
ylabel={Character accuracy},
x grid style={white!69.0196078431373!black},
xmin=2.6, xmax=11.4,
xtick style={color=black},
y grid style={white!69.0196078431373!black},
ymin=0, ymax=1.01,
ytick style={color=black},
xmajorgrids,
ymajorgrids,
every axis plot/.append style={thick}
]
\addplot [semithick, color0,  line width=1.5pt]
table {%
3 0.791385714285714
5 0.296014285714286
7 0.129571428571429
9 0.0959142857142857
11 0.0870857142857142
};
\addplot [semithick, color1,  line width=1.5pt]
table {%
3 0.949871428571429
5 0.359871428571429
7 0.216985714285714
9 0.175585714285714
11 0.151528571428571
};
\addplot [semithick, color2,  line width=1.5pt]
table {%
3 0.987185714285714
5 0.664614285714286
7 0.499585714285714
9 0.343828571428571
11 0.252971428571429
};
\addplot [semithick, color3,  line width=1.5pt]
table {%
3 0.956785714285714
5 0.419114285714286
7 0.209542857142857
9 0.111414285714286
11 0.0903
};
\end{axis}

\end{tikzpicture}

%% file: tex/noise-uncertainty-sp-AUC-pr-5.tex
\begin{tikzpicture}

\definecolor{color0}{rgb}{0.00392156862745098, 0.45098039215686275, 0.6980392156862745}
\definecolor{color1}{rgb}{0.8705882352941177, 0.5607843137254902, 0.019607843137254}
\definecolor{color2}{rgb}{0.00784313725490196, 0.6196078431372549, 0.45098039215686275}
\definecolor{color3}{rgb}{0.8352941176470589, 0.3686274509803922, 0.0}

\begin{axis}[
legend cell align={left},
legend style={
	fill opacity=0.8,
	draw opacity=1,
	text opacity=1,
	at={(0.5,1.22)},
	anchor=north,
	draw=white!80!black,
	legend columns=2
},
width=0.951\figurewidth,
height=\figureheight,
at={(0\figurewidth,0\figureheight)},
scale only axis,
ylabel near ticks,
xlabel near ticks,
log basis x={10},
tick align=outside,
tick pos=left,
xlabel={$p$},
ylabel={AUC},
x grid style={white!69.0196078431373!black},
xmin=6.60536773049804e-05, xmax=0.605568102065137,
xmode=log,
xtick style={color=black},
y grid style={white!69.0196078431373!black},
ymin=0.4, ymax=1.01,
ytick style={color=black},
xmajorgrids,
ymajorgrids,
every axis plot/.append style={thick}
]
\addplot [semithick, color0,  line width=1.5pt]
table {%
0.0001 0.541666666666667
0.001 0.530607769423559
0.005 0.571958035604036
0.01 0.519264990857084
0.025 0.482015069587141
0.05 0.464803900952019
0.075 0.451489628447306
0.1 0.449817941516172
0.15 0.440933794665668
0.2 0.44163443531592
0.25 0.445009431305246
0.3 0.452311326456677
0.35 0.466940858172857
0.4 0.477824918412477
};
\addplot [semithick, color1,  line width=1.5pt]
table {%
0.0001 1
0.001 0.988804430427566
0.005 0.977254923696231
0.01 0.969314642673421
0.025 0.961437879321596
0.05 0.910935008487998
0.075 0.846607965937636
0.1 0.811050529168362
0.15 0.717049574464405
0.2 0.647212034627115
0.25 0.599081595514241
0.3 0.562156240220569
0.35 0.528089177027863
0.4 0.510716614388976
};
\addplot [semithick, color2,  line width=1.5pt]
table {%
0.0001 0.951007200162504
0.001 0.890430327286241
0.005 0.910225665564476
0.01 0.941603903476612
0.025 0.896617687791347
0.05 0.852591996701236
0.075 0.779771702131236
0.1 0.717448854068614
0.15 0.603832074238345
0.2 0.525163470505922
0.25 0.47198835289012
0.3 0.437247416697318
0.35 0.415852626089801
0.4 0.404131319000193
};
\addplot [semithick, color3,  line width=1.5pt]
table {%
0.001 0.983119658119658
0.005 0.986980952854433
0.01 0.954040373815202
0.025 0.93215404599885
0.05 0.873916686918891
0.075 0.810306803900545
0.1 0.732973885097967
0.15 0.6227690815516
0.2 0.546694191749642
0.25 0.502294788183645
0.3 0.477782672178316
0.35 0.463877448245009
0.4 0.464017859048993
};
\end{axis}

\end{tikzpicture}

%% file: tex/noise-uncertainty-gaussian-AUC-pr-5.tex
\begin{tikzpicture}

\definecolor{color0}{rgb}{0.00392156862745098, 0.45098039215686275, 0.6980392156862745}
\definecolor{color1}{rgb}{0.8705882352941177, 0.5607843137254902, 0.019607843137254}
\definecolor{color2}{rgb}{0.00784313725490196, 0.6196078431372549, 0.45098039215686275}
\definecolor{color3}{rgb}{0.8352941176470589, 0.3686274509803922, 0.0}

\begin{axis}[
legend cell align={left},
legend style={
	fill opacity=0.8,
	draw opacity=1,
	text opacity=1,
	at={(0.5,1.22)},
	anchor=north,
	draw=white!80!black,
	legend columns=2
},
width=0.951\figurewidth,
height=\figureheight,
at={(0\figurewidth,0\figureheight)},
scale only axis,
ylabel near ticks,
xlabel near ticks,
log basis x={10},
tick align=outside,
tick pos=left,
xlabel={$\sigma$},
ylabel={AUC},
x grid style={white!69.0196078431373!black},
xmin=6.60536773049804e-05, xmax=0.605568102065137,
xmode=log,
xtick style={color=black},
y grid style={white!69.0196078431373!black},
ymin=0.4, ymax=1.01,
ytick style={color=black},
xmajorgrids,
ymajorgrids,
every axis plot/.append style={thick}
]
\addplot [semithick, color0,  line width=1.5pt]
table {%
0.05 0.75
0.075 0.458419599723948
0.1 0.470384234183622
0.15 0.438444298994644
0.2 0.434086261533028
0.25 0.428383197641816
0.3 0.435197146157734
0.35 0.441110918027392
0.4 0.452548772102613
};
\addplot [semithick, color1,  line width=1.5pt]
table {%
0.05 1
0.075 0.989894067066053
0.1 0.981731698785911
0.15 0.830236625703813
0.2 0.67597346068715
0.25 0.573266329810601
0.3 0.523211737362012
0.35 0.493883712287503
0.4 0.488105775457146
};
\addplot [semithick, color2,  line width=1.5pt]
table {%
0.0001 0.92254370094269
0.001 0.951499145286636
0.005 0.952868359111339
0.01 0.896681404628708
0.025 0.886830037654828
0.05 0.900258835976804
0.075 0.922190095034102
0.1 0.82342710657682
0.15 0.633289850669438
0.2 0.479232271157779
0.25 0.414229258129482
0.3 0.39566068232748
0.35 0.397892062806753
0.4 0.405807902964963
};
\addplot [semithick, color3,  line width=1.5pt]
table {%
0.075 1
0.1 0.965411486382987
0.15 0.880397694796246
0.2 0.72067375150233
0.25 0.607399939063136
0.3 0.532204651488069
0.35 0.492610539621563
0.4 0.46804649134401
};
\end{axis}

\end{tikzpicture}

%% file: tex/blur-uncertainty-horizontal-AUC-pr-5.tex
\begin{tikzpicture}

\definecolor{color0}{rgb}{0.00392156862745098, 0.45098039215686275, 0.6980392156862745}
\definecolor{color1}{rgb}{0.8705882352941177, 0.5607843137254902, 0.019607843137254}
\definecolor{color2}{rgb}{0.00784313725490196, 0.6196078431372549, 0.45098039215686275}
\definecolor{color3}{rgb}{0.8352941176470589, 0.3686274509803922, 0.0}
\begin{axis}[
legend cell align={left},
legend style={
	fill opacity=0.8,
	draw opacity=1,
	text opacity=1,
	at={(0.5,1.22)},
	anchor=north,
	draw=white!80!black,
	legend columns=2
},
width=0.951\figurewidth,
height=\figureheight,
at={(0\figurewidth,0\figureheight)},
scale only axis,
tick align=outside,
tick pos=left,
x grid style={white!69.0196078431373!black},
xmin=2.6, xmax=11.4,
xtick style={color=black},
y grid style={white!69.0196078431373!black},
ymin=0.37, ymax=0.855,
ytick style={color=black},
ylabel near ticks,
xlabel={$k$},
ylabel={AUC},
xlabel near ticks,
xmajorgrids,
ymajorgrids,
every axis plot/.append style={thick}
]
\addplot [semithick, color0,  line width=1.5pt]
table {%
3 0.509833350313243
5 0.561760041570653
7 0.538351411260934
9 0.521999689614463
11 0.518417200801777
};
\addplot [semithick, color1,  line width=1.5pt]
table {%
3 0.738138427159396
5 0.519152935536174
7 0.501393418510206
9 0.496088136104329
11 0.497903477889093
};
\addplot [semithick, color2,  line width=1.5pt]
table {%
3 0.780016536583183
5 0.449614587145411
7 0.387625024516493
9 0.385263659451928
11 0.389490089191011
};
\addplot [semithick, color3,  line width=1.5pt]
table {%
3 0.697772848723541
5 0.545430263566076
7 0.478211833355288
9 0.463155611981909
11 0.470074069616937
};
\end{axis}

\end{tikzpicture}

%% file: tex/blur-uncertainty-vertical-AUC-pr-5.tex
\begin{tikzpicture}

\definecolor{color0}{rgb}{0.00392156862745098, 0.45098039215686275, 0.6980392156862745}
\definecolor{color1}{rgb}{0.8705882352941177, 0.5607843137254902, 0.019607843137254}
\definecolor{color2}{rgb}{0.00784313725490196, 0.6196078431372549, 0.45098039215686275}
\definecolor{color3}{rgb}{0.8352941176470589, 0.3686274509803922, 0.0}

\begin{axis}[
legend cell align={left},
legend style={
	fill opacity=0.8,
	draw opacity=1,
	text opacity=1,
	at={(0.5,1.22)},
	anchor=north,
	draw=white!80!black,
	legend columns=2
},
width=0.951\figurewidth,
height=\figureheight,
at={(0\figurewidth,0\figureheight)},
scale only axis,
ylabel near ticks,
xlabel near ticks,
tick align=outside,
tick pos=left,
x grid style={white!69.0196078431373!black},
xmin=2.6, xmax=11.4,
xlabel={$k$},
ylabel={AUC},
xtick style={color=black},
y grid style={white!69.0196078431373!black},
ymin=0.37, ymax=0.855,
ytick style={color=black},
xmajorgrids,
ymajorgrids,
every axis plot/.append style={thick}
]
\addplot [semithick, color0,  line width=1.5pt]
table {%
3 0.473882213248017
5 0.504816436775801
7 0.561384057089844
9 0.561623343078881
11 0.558450768970248
};
\addplot [semithick, color1,  line width=1.5pt]
table {%
3 0.870978996888416
5 0.664784594523035
7 0.614703014755553
9 0.584375441334297
11 0.582380908573224
};
\addplot [semithick, color2,  line width=1.5pt]
table {%
3 0.85043068912401
5 0.673280720805476
7 0.585099230308296
9 0.570850872645832
11 0.581176547488697
};
\addplot [semithick, color3,  line width=1.5pt]
table {%
3 0.780972604506919
5 0.583034700770406
7 0.505757012787494
9 0.562934646928832
11 0.554553651488441
};
\end{axis}

\end{tikzpicture}

%% file: tex/noise-sp-median-001.tex
\begin{tikzpicture}

	\definecolor{color0}{rgb}{0.8705882352941177, 0.5607843137254902, 0.019607843137254}
	\definecolor{color1}{rgb}{0.00784313725490196, 0.6196078431372549, 0.45098039215686275}

\begin{axis}[
legend cell align={left},
legend cell align={left},legend style={
	fill opacity=0.8,
	draw opacity=1,
	text opacity=1,
	at={(0.5,1.22)},
	anchor=north,
	draw=white!80!black,
	legend columns=2
},
width=0.951\figurewidth,
height=\figureheight,
at={(0\figurewidth,0\figureheight)},
scale only axis,
ylabel near ticks,
xlabel near ticks,
xmajorgrids,
ymajorgrids,
every axis plot/.append style={thick},
tick align=outside,
tick pos=left,
xlabel={Inference steps},
ylabel={Predictive uncertainty},
x grid style={white!69.0196078431373!black},
xmin=2.75, xmax=52.25,
xtick style={color=black},
y grid style={white!69.0196078431373!black},
ymin=-0.01, ymax=0.463297468423843,
ytick style={color=black}
]
\addplot [semithick, color0,  line width=1.5pt]
table {%
5 9.80738241196377e-06
10 2.94950132229133e-05
20 7.4132298323093e-05
30 0.0001220171834575
40 0.00017254492559
50 0.0002243927010567
};
\addplot [semithick, color0,dashed,  line width=1.5pt]
table {%
5 0.390683740377426
10 0.415164858102798
20 0.419946983456612
30 0.417575031518936
40 0.413143709301948
50 0.419872939586639
};
\addplot [semithick, color1,  line width=1.5pt]
table {%
5 0.0113162528723478
10 0.0259775780141353
20 0.0502830520272254
30 0.0726633258163929
40 0.0868262946605682
50 0.0975436754524707
};
\addplot [semithick, color1,dashed,  line width=1.5pt]
table {%
5 0.387163534760475
10 0.397689417004585
20 0.427830666303635
30 0.42482590675354
40 0.450416803359985
50 0.444288164377213
};
\end{axis}

\end{tikzpicture}

%% file: tex/noise-gaussian-median-02.tex
\begin{tikzpicture}

	\definecolor{color0}{rgb}{0.8705882352941177, 0.5607843137254902, 0.019607843137254}
	\definecolor{color1}{rgb}{0.00784313725490196, 0.6196078431372549, 0.45098039215686275}

\begin{axis}[
legend cell align={left},
legend cell align={left},legend style={
	fill opacity=0.8,
	draw opacity=1,
	text opacity=1,
	at={(0.5,1.22)},
	anchor=north,
	draw=white!80!black,
	legend columns=2
},
width=0.951\figurewidth,
height=\figureheight,
at={(0\figurewidth,0\figureheight)},
scale only axis,
ylabel near ticks,
xlabel near ticks,
xmajorgrids,
ymajorgrids,
xlabel={Inference steps},
ylabel={Predictive uncertainty},
every axis plot/.append style={thick},
tick align=outside,
tick pos=left,
x grid style={white!69.0196078431373!black},
xmin=2.75, xmax=52.25,
xtick style={color=black},
y grid style={white!69.0196078431373!black},
ymin=-0.01, ymax=0.463297468423843,
ytick style={color=black}
]
\addplot [semithick, color0,  line width=1.5pt]
table {%
5 0.125235050916672
10 0.280070513486862
20 0.274491339921951
30 0.275529593229294
40 0.275581747293472
50 0.278974920511246
};
\addplot [semithick, color0,dashed,  line width=1.5pt]
table {%
5 0.391688615083694
10 0.395398139953613
20 0.399580210447311
30 0.399243891239166
40 0.397970944643021
50 0.399544507265091
};
\addplot [semithick, color1,  line width=1.5pt]
table {%
5 0.326115429401398
10 0.335048764944076
20 0.341202989220619
30 0.343018889427185
40 0.344137042760849
50 0.344935089349747
};
\addplot [semithick, color1,dashed,  line width=1.5pt]
table {%
5 0.319512844085693
10 0.306427508592606
20 0.301588118076324
30 0.299572736024857
40 0.297555476427078
50 0.297985345125198
};
\end{axis}

\end{tikzpicture}

%% file: tex/blur-horizontal-median-5.tex
\begin{tikzpicture}

  \definecolor{color0}{rgb}{0.8705882352941177, 0.5607843137254902, 0.019607843137254}
  \definecolor{color1}{rgb}{0.00784313725490196, 0.6196078431372549, 0.45098039215686275}

\begin{axis}[
legend cell align={left},
legend cell align={left},legend style={
	fill opacity=0.8,
	draw opacity=1,
	text opacity=1,
	at={(0.5,1.22)},
	anchor=north,
	draw=white!80!black,
	legend columns=2
},
width=0.951\figurewidth,
height=\figureheight,
at={(0\figurewidth,0\figureheight)},
scale only axis,
ylabel near ticks,
xlabel near ticks,
xmajorgrids,
ymajorgrids,
xlabel={Inference steps},
ylabel={Predictive uncertainty},
every axis plot/.append style={thick},
tick align=outside,
tick pos=left,
x grid style={white!69.0196078431373!black},
xmin=2.75, xmax=52.25,
xtick style={color=black},
y grid style={white!69.0196078431373!black},
ymin=-0.01, ymax=0.463297468423843,
ytick style={color=black}
]
\addplot [semithick, color0,  line width=1.5pt]
table {%
5 0.323813706636429
10 0.338919520378113
20 0.344577431678772
30 0.349384218454361
40 0.349850386381149
50 0.3524529337883
};
\addplot [semithick, color0,dashed,  line width=1.5pt]
table {%
5 0.338287860155106
10 0.348478257656097
20 0.353557795286179
30 0.355264753103256
40 0.355914950370789
50 0.356729105114937
};
\addplot [semithick, color1,  line width=1.5pt]
table {%
5 0.269231632351875
10 0.288027137517929
20 0.295493930578232
30 0.298901170492172
40 0.300606682896614
50 0.301124364137649
};
\addplot [semithick, color1,dashed,  line width=1.5pt]
table {%
5 0.240232162177563
10 0.239317864179611
20 0.239994272589684
30 0.24088180065155
40 0.240982316434383
50 0.240413874387741
};
\end{axis}

\end{tikzpicture}

%% file: tex/noise-sp-spread-001.tex
\begin{tikzpicture}
	\definecolor{color0}{rgb}{0.8705882352941177, 0.5607843137254902, 0.019607843137254}
	\definecolor{color1}{rgb}{0.00784313725490196, 0.6196078431372549, 0.45098039215686275}

\begin{axis}[
legend cell align={left},
legend cell align={left},legend style={
	fill opacity=0.8,
	draw opacity=1,
	text opacity=1,
	at={(0.5,1.22)},
	anchor=north,
	draw=white!80!black,
	legend columns=2
},
width=0.951\figurewidth,
height=\figureheight,
at={(0\figurewidth,0\figureheight)},
scale only axis,
ylabel near ticks,
xlabel near ticks,
xmajorgrids,
ymajorgrids,
every axis plot/.append style={thick},
tick align=outside,
tick pos=left,
xlabel={Inference steps},
ylabel={Predictive uncertainty},
x grid style={white!69.0196078431373!black},
xmin=2.75, xmax=52.25,
xtick style={color=black},
y grid style={white!69.0196078431373!black},
ymin=-0.00810851163450941, ymax=0.172687262565978,
ytick style={color=black}
]
\addplot [semithick, color0,  line width=1.5pt]
table {%
5 0.000109478101876384
10 0.000293389930561679
20 0.000732631399387112
30 0.00124709298984264
40 0.00182574112068315
50 0.00200372406288806
};
\addplot [semithick, color0,dashed,  line width=1.5pt]
table {%
5 0.120383054018021
10 0.0595782995223999
20 0.0336119830608368
30 0.0540729165077209
40 0.0625394582748413
50 0.0578141957521438
};
\addplot [semithick, color1,  line width=1.5pt]
table {%
5 0.063225666352082
10 0.114103918196633
20 0.164469272829592
30 0.159093523863703
40 0.142036768607795
50 0.135640283115208
};
\addplot [semithick, color1,dashed,  line width=1.5pt]
table {%
5 0.0978276059031487
10 0.0873624533414841
20 0.0576340258121491
30 0.0264912918210029
40 0.00502461194992071
50 0.0582465305924416
};
\end{axis}

\end{tikzpicture}

%% file: tex/noise-gaussian-spread-02.tex
\begin{tikzpicture}

	\definecolor{color0}{rgb}{0.8705882352941177, 0.5607843137254902, 0.019607843137254}
	\definecolor{color1}{rgb}{0.00784313725490196, 0.6196078431372549, 0.45098039215686275}

\begin{axis}[
legend cell align={left},
legend cell align={left},legend style={
	fill opacity=0.8,
	draw opacity=1,
	text opacity=1,
	at={(0.5,1.22)},
	anchor=north,
	draw=white!80!black,
	legend columns=2
},
width=0.951\figurewidth,
height=\figureheight,
at={(0\figurewidth,0\figureheight)},
scale only axis,
ylabel near ticks,
xlabel near ticks,
xmajorgrids,
ymajorgrids,
xlabel={Inference steps},
ylabel={Predictive uncertainty},
every axis plot/.append style={thick},
tick align=outside,
tick pos=left,
x grid style={white!69.0196078431373!black},
xmin=2.75, xmax=52.25,
xtick style={color=black},
y grid style={white!69.0196078431373!black},
ymin=0.0389363015943673, ymax=0.40469488001545,
ytick style={color=black}
]
\addplot [semithick, color0,  line width=1.5pt]
table {%
5 0.388069490087219
10 0.381478345254436
20 0.336615525186062
30 0.293670151382685
40 0.261965192854404
50 0.254852861166
};
\addplot [semithick, color0,dashed,  line width=1.5pt]
table {%
5 0.0905845314264298
10 0.0687370300292969
20 0.0617009699344635
30 0.0598877370357513
40 0.056280106306076
50 0.0555616915225983
};
\addplot [semithick, color1,  line width=1.5pt]
table {%
5 0.185149535536766
10 0.109341561794281
20 0.0995305106043816
30 0.0938608050346375
40 0.09230937063694
50 0.0918828025460244
};
\addplot [semithick, color1,dashed,  line width=1.5pt]
table {%
5 0.113587245345116
10 0.102945178747177
20 0.099787849932909
30 0.0970963537693024
40 0.0978616029024124
50 0.103910505771637
};
\end{axis}

\end{tikzpicture}

%% file: tex/blur-horizontal-spread-5.tex
\begin{tikzpicture}

	\definecolor{color0}{rgb}{0.8705882352941177, 0.5607843137254902, 0.019607843137254}
	\definecolor{color1}{rgb}{0.00784313725490196, 0.6196078431372549, 0.45098039215686275}

\begin{axis}[
legend cell align={left},legend style={
	fill opacity=0.8,
	draw opacity=1,
	text opacity=1,
	at={(0.5,1.22)},
	anchor=north,
	draw=white!80!black,
	legend columns=2
},
width=0.951\figurewidth,
height=\figureheight,
at={(0\figurewidth,0\figureheight)},
scale only axis,
ylabel near ticks,
xlabel near ticks,
xmajorgrids,
ymajorgrids,
xlabel={Inference steps},
ylabel={Predictive uncertainty},
every axis plot/.append style={thick},
tick align=outside,
tick pos=left,
x grid style={white!69.0196078431373!black},
xmin=2.75, xmax=52.25,
xtick style={color=black},
y grid style={white!69.0196078431373!black},
ymin=0.0493192465975881, ymax=0.225576745532453,
ytick style={color=black}
]
\addplot [semithick, color0,  line width=1.5pt]
table {%
5 0.217565041035414
10 0.124774500727654
20 0.112775214016438
30 0.0967554897069931
40 0.103357955813408
50 0.101605825126171
};
\addplot [semithick, color0, dashed,  line width=1.5pt]
table {%
5 0.128371998667717
10 0.0903327167034149
20 0.0679761692881584
30 0.063868522644043
40 0.0612258315086365
50 0.0573309510946274
};
\addplot [semithick, color1,  line width=1.5pt]
table {%
5 0.164357766509056
10 0.109407536685467
20 0.0927699655294418
30 0.089335672557354
40 0.0856227576732635
50 0.0843984782695771
};
\addplot [semithick, color1, dashed,  line width=1.5pt]
table {%
5 0.137362152338028
10 0.108206212520599
20 0.098095029592514
30 0.0977616459131241
40 0.095987293869257
50 0.0953693091869355
};
\end{axis}

\end{tikzpicture}

%% file: tex/noise-sp-median.tex
\begin{tikzpicture}

	\definecolor{color0}{rgb}{0.8705882352941177, 0.5607843137254902, 0.019607843137254}
	\definecolor{color1}{rgb}{0.00784313725490196, 0.6196078431372549, 0.45098039215686275}

\begin{axis}[
  legend cell align={left},
  legend cell align={left},legend style={
    fill opacity=0.8,
    draw opacity=1,
    text opacity=1,
    at={(0.5,1.22)},
    anchor=north,
    draw=white!80!black,
    legend columns=2
  },
  width=0.951\figurewidth,
  height=\figureheight,
  at={(0\figurewidth,0\figureheight)},
  scale only axis,
  ylabel near ticks,
  xlabel near ticks,
  xmajorgrids,
  ymajorgrids,
  every axis plot/.append style={thick},
log basis x={10},
tick align=outside,
tick pos=left,
ylabel={Predictive uncertainty},
xlabel={$p$},
x grid style={white!69.0196078431373!black},
xmin=6.60536773049804e-05, xmax=0.605568102065137,
xmode=log,
xtick style={color=black},
y grid style={white!69.0196078431373!black},
ymin=-0.0201931230723858, ymax=0.424055584520102,
ytick style={color=black}
]
\addplot [semithick, color0,  line width=1.5pt]
table {%
0.0001 2.22786566155264e-05
0.001 2.12991344596958e-05
0.005 2.5908316274581e-05
0.01 2.94950132229133e-05
0.025 5.14921921421774e-05
0.05 0.0001371523540001
0.075 0.000596987287281
0.1 0.0030747654382139
0.15 0.126745879650116
0.2 0.29218065738678
0.25 0.340761214494705
0.3 0.367802500724792
0.35 0.381180912256241
0.4 0.389904946088791
};
\addplot [semithick, color0,dashed,  line width=1.5pt]
table {%
0.0001 0
0.001 0.380571275949478
0.005 0.411049798130989
0.01 0.415164858102798
0.025 0.419483929872513
0.05 0.40728448331356
0.075 0.410774886608124
0.1 0.401338458061218
0.15 0.401887863874435
0.2 0.39754393696785
0.25 0.391764730215073
0.3 0.392398059368133
0.35 0.391163617372513
0.4 0.389104276895523
};
\addplot [semithick, color1,  line width=1.5pt]
table {%
0.0001 0.0199288073927164
0.001 0.0213096160441637
0.005 0.0233975630253553
0.01 0.0259775780141353
0.025 0.0383983980864286
0.05 0.0810783132910728
0.075 0.155634194612503
0.1 0.247406609356403
0.15 0.298391968011856
0.2 0.331691950559616
0.25 0.343791574239731
0.3 0.343881174921989
0.35 0.34263613820076
0.4 0.33801606297493
};
\addplot [semithick, color1,dashed,  line width=1.5pt]
table {%
0.0001 0.322537034749985
0.001 0.337631464004517
0.005 0.402419120073318
0.01 0.397689417004585
0.025 0.414370715618133
0.05 0.401922911405563
0.075 0.388929486274719
0.1 0.378085106611252
0.15 0.35854297876358
0.2 0.325540065765381
0.25 0.307373657822609
0.3 0.287022113800049
0.35 0.271247029304504
0.4 0.256839126348495
};
\end{axis}

\end{tikzpicture}

%% file: tex/noise-gaussian-median.tex
\begin{tikzpicture}

	\definecolor{color0}{rgb}{0.8705882352941177, 0.5607843137254902, 0.019607843137254}
	\definecolor{color1}{rgb}{0.00784313725490196, 0.6196078431372549, 0.45098039215686275}s
  
  \begin{axis}[
  legend cell align={left},
  legend cell align={left},legend style={
    fill opacity=0.8,
    draw opacity=1,
    text opacity=1,
    at={(0.5,1.22)},
    anchor=north,
    draw=white!80!black,
    legend columns=2
  },
  width=0.951\figurewidth,
  height=\figureheight,
  at={(0\figurewidth,0\figureheight)},
  scale only axis,
  ylabel near ticks,
  xlabel near ticks,
  xmajorgrids,
  ylabel={Predictive uncertainty},
xlabel={$\sigma$},
  ymajorgrids,
  every axis plot/.append style={thick},
log basis x={10},
tick align=outside,
tick pos=left,
x grid style={white!69.0196078431373!black},
xmin=6.60536773049804e-05, xmax=0.605568102065137,
xmode=log,
xtick style={color=black},
y grid style={white!69.0196078431373!black},
ymin=-0.0201931230723858, ymax=0.424055584520102,
ytick style={color=black}
]
\addplot [semithick, color0,  line width=1.5pt]
table {%
0.0001 2.23273827941739e-05
0.001 2.09198969969293e-05
0.005 2.07113480428234e-05
0.01 2.00387403310742e-05
0.025 2.11068454518681e-05
0.05 2.54641545325285e-05
0.075 3.8629299524473e-05
0.1 8.14793638710398e-05
0.15 0.0054394216276705
0.2 0.280070513486862
0.25 0.356435000896454
0.3 0.381203457713127
0.35 0.388661712408066
0.4 0.391014516353607
};
\addplot [semithick, color0,dashed,  line width=1.5pt]
table {%
0.0001 0
0.001 0
0.005 0
0.01 0
0.025 0
0.05 0
0.075 0.465960502624512
0.1 0.415165454149246
0.15 0.404392212629318
0.2 0.395398139953613
0.25 0.390211522579193
0.3 0.386505246162414
0.35 0.385862499475479
0.4 0.383549481630325
};
\addplot [semithick, color1,  line width=1.5pt]
table {%
0.0001 0.0203727521002292
0.001 0.0201477631926536
0.005 0.0201562847942113
0.01 0.0198977682739496
0.025 0.0212792102247476
0.05 0.0262168291956186
0.075 0.0440995674580335
0.1 0.101119186729193
0.15 0.291285842657089
0.2 0.335048764944076
0.25 0.339054435491562
0.3 0.324666574597359
0.35 0.306427925825119
0.4 0.28448361158371
};
\addplot [semithick, color1,dashed,  line width=1.5pt]
table {%
0.0001 0.38130995631218
0.001 0.333460450172424
0.005 0.333141922950745
0.01 0.341718271374702
0.025 0.340507596731186
0.05 0.357300907373428
0.075 0.356529206037521
0.1 0.355822294950485
0.15 0.343835204839706
0.2 0.306427508592606
0.25 0.277383178472519
0.3 0.250569507479668
0.35 0.234079338610172
0.4 0.223087280988693
};
\end{axis}

\end{tikzpicture}

%% file: tex/blur-horizontal-median.tex
\begin{tikzpicture}

	\definecolor{color0}{rgb}{0.8705882352941177, 0.5607843137254902, 0.019607843137254}
	\definecolor{color1}{rgb}{0.00784313725490196, 0.6196078431372549, 0.45098039215686275}
  
  \begin{axis}[
  legend cell align={left},
  legend cell align={left},legend style={
    fill opacity=0.8,
    draw opacity=1,
    text opacity=1,
    at={(0.5,1.22)},
    anchor=north,
    draw=white!80!black,
    legend columns=2
  },
  width=0.951\figurewidth,
  height=\figureheight,
  at={(0\figurewidth,0\figureheight)},
  scale only axis,
  ylabel near ticks,
  xlabel near ticks,
  xmajorgrids,
  ylabel={Predictive uncertainty},
xlabel={$k$},
  ymajorgrids,
  every axis plot/.append style={thick},
tick align=outside,
tick pos=left,
x grid style={white!69.0196078431373!black},
xmin=2.6, xmax=11.4,
xtick style={color=black},
y grid style={white!69.0196078431373!black},
ymin=-0.0201931230723858, ymax=0.424055584520102,
ytick style={color=black}
]
\addplot [semithick, color0,  line width=1.5pt]
table {%
3 0.0380711629986763
5 0.338919520378113
7 0.337646722793579
9 0.333947226405144
11 0.329714477062225
};
\addplot [semithick, color0,dashed,  line width=1.5pt]
table {%
3 0.385148167610168
5 0.348478257656097
7 0.333611786365509
9 0.334326952695847
11 0.333549559116364
};
\addplot [semithick, color1,  line width=1.5pt]
table {%
3 0.133214220404625
5 0.288027137517929
7 0.264784634113312
9 0.245230689644814
11 0.226805374026298
};
\addplot [semithick, color1,dashed,  line width=1.5pt]
table {%
3 0.355099886655807
5 0.239317864179611
7 0.188915207982063
9 0.169382035732269
11 0.158945322036743
};
\end{axis}

\end{tikzpicture}

%% file: tex/blur-vertical-median.tex
\begin{tikzpicture}

	\definecolor{color0}{rgb}{0.8705882352941177, 0.5607843137254902, 0.019607843137254}
	\definecolor{color1}{rgb}{0.00784313725490196, 0.6196078431372549, 0.45098039215686275}
  
  \begin{axis}[
  legend cell align={left},
  legend cell align={left},legend style={
    fill opacity=0.8,
    draw opacity=1,
    text opacity=1,
    at={(0.5,1.22)},
    anchor=north,
    draw=white!80!black,
    legend columns=2
  },
  width=0.951\figurewidth,
  height=\figureheight,
  at={(0\figurewidth,0\figureheight)},
  scale only axis,
  ylabel near ticks,
  xlabel near ticks,
  ylabel={Predictive uncertainty},
xlabel={$k$},
  xmajorgrids,
  ymajorgrids,
  every axis plot/.append style={thick},
tick align=outside,
tick pos=left,
x grid style={white!69.0196078431373!black},
xmin=2.6, xmax=11.4,
xtick style={color=black},
y grid style={white!69.0196078431373!black},
ymin=-0.0201931230723858, ymax=0.424055584520102,
ytick style={color=black}
]
\addplot [semithick, color0,  line width=1.5pt]
table {%
3 0.0006112350383773
5 0.179757684469223
7 0.275417372584343
9 0.287867069244385
11 0.292083188891411
};
\addplot [semithick, color0,dashed,  line width=1.5pt]
table {%
3 0.403934627771378
5 0.36896476149559
7 0.351861298084259
9 0.345608681440353
11 0.340051934123039
};
\addplot [semithick, color1,  line width=1.5pt]
table {%
3 0.0842932909727096
5 0.245112910866737
7 0.290242344141006
9 0.284065932035446
11 0.271065384149551
};
\addplot [semithick, color1,dashed,  line width=1.5pt]
table {%
3 0.353863343596458
5 0.329844385385513
7 0.320451721549034
9 0.312998592853546
11 0.31155601143837
};
\end{axis}

\end{tikzpicture}

%% file: paper.bbl
\begin{thebibliography}{10}
\providecommand{\url}[1]{#1}
\csname url@samestyle\endcsname
\providecommand{\newblock}{\relax}
\providecommand{\bibinfo}[2]{#2}
\providecommand{\BIBentrySTDinterwordspacing}{\spaceskip=0pt\relax}
\providecommand{\BIBentryALTinterwordstretchfactor}{4}
\providecommand{\BIBentryALTinterwordspacing}{\spaceskip=\fontdimen2\font plus
\BIBentryALTinterwordstretchfactor\fontdimen3\font minus
  \fontdimen4\font\relax}
\providecommand{\BIBforeignlanguage}[2]{{%
\expandafter\ifx\csname l@#1\endcsname\relax
\typeout{** WARNING: IEEEtran.bst: No hyphenation pattern has been}%
\typeout{** loaded for the language `#1'. Using the pattern for}%
\typeout{** the default language instead.}%
\else
\language=\csname l@#1\endcsname
\fi
#2}}
\providecommand{\BIBdecl}{\relax}
\BIBdecl

\bibitem{Silva2018}
S.~M. Silva and C.~R. Jung, ``License plate detection and recognition in
  unconstrained scenarios,'' in \emph{Proceedings of the European conference on
  computer vision (ECCV)}, 2018, pp. 580--596.

\bibitem{Zhang2020}
L.~Zhang, P.~Wang, H.~Li, Z.~Li, C.~Shen, and Y.~Zhang, ``A robust attentional
  framework for license plate recognition in the wild,'' \emph{IEEE
  Transactions on Intelligent Transportation Systems}, pp. 1--10, 2020.

\bibitem{Kaiser2021}
P.~Kaiser, F.~Schirrmacher, B.~Lorch, and C.~Riess, ``{Learning to Decipher
  License Plates in Severely Degraded Images},'' in \emph{Pattern Recognition.
  ICPR International Workshops and Challenges}.\hskip 1em plus 0.5em minus
  0.4em\relax Springer International Publishing, 2021, pp. 544--559.

\bibitem{Kendall2017}
A.~Kendall and Y.~Gal, ``What uncertainties do we need in bayesian deep
  learning for computer vision?'' in \emph{Advances in Neural Information
  Processing Systems 30: Annual Conference on Neural Information Processing
  Systems 2017, December 4-9, 2017, Long Beach, CA, {USA}}, I.~Guyon, U.~von
  Luxburg, S.~Bengio, H.~M. Wallach, R.~Fergus, S.~V.~N. Vishwanathan, and
  R.~Garnett, Eds., 2017, pp. 5574--5584.

\bibitem{Snoek2019}
J.~Snoek, Y.~Ovadia, E.~Fertig, B.~Lakshminarayanan, S.~Nowozin, D.~Sculley,
  J.~Dillon, J.~Ren, and Z.~Nado, ``Can you trust your model's uncertainty?
  evaluating predictive uncertainty under dataset shift,'' in \emph{Advances in
  Neural Information Processing Systems}, 2019, pp. 13\,969--13\,980.

\bibitem{Maier2020}
A.~Maier, B.~Lorch, and C.~Riess, ``Toward reliable models for authenticating
  multimedia content: Detecting resampling artifacts with bayesian neural
  networks,'' in \emph{2020 IEEE International Conference on Image Processing
  (ICIP)}.\hskip 1em plus 0.5em minus 0.4em\relax IEEE, 2020, pp. 1251--1255.

\bibitem{Ovadia2019}
Y.~Ovadia, E.~Fertig, J.~Ren, Z.~Nado, D.~Sculley, S.~Nowozin, J.~Dillon,
  B.~Lakshminarayanan, and J.~Snoek, ``Can you trust your model's uncertainty?
  evaluating predictive uncertainty under dataset shift,'' \emph{Advances in
  Neural Information Processing Systems}, vol.~32, pp. 13\,991--14\,002, 2019.

\bibitem{Lakshminarayanan2017}
B.~Lakshminarayanan, A.~Pritzel, and C.~Blundell, ``Simple and scalable
  predictive uncertainty estimation using deep ensembles,'' in \emph{Advances
  in Neural Information Processing Systems 30: Annual Conference on Neural
  Information Processing Systems 2017, December 4-9, 2017, Long Beach, CA,
  {USA}}, I.~Guyon, U.~von Luxburg, S.~Bengio, H.~M. Wallach, R.~Fergus,
  S.~V.~N. Vishwanathan, and R.~Garnett, Eds., 2017, pp. 6402--6413.

\bibitem{Gal2016}
Y.~Gal and Z.~Ghahramani, ``Dropout as a bayesian approximation: Representing
  model uncertainty in deep learning,'' in \emph{Proceedings of The 33rd
  International Conference on Machine Learning}, ser. Proceedings of Machine
  Learning Research, M.~F. Balcan and K.~Q. Weinberger, Eds., vol.~48.\hskip
  1em plus 0.5em minus 0.4em\relax New York, New York, USA: PMLR, 20--22 Jun
  2016, pp. 1050--1059.

\bibitem{Wen2020}
\BIBentryALTinterwordspacing
Y.~Wen, D.~Tran, and J.~Ba, ``Batchensemble: an alternative approach to
  efficient ensemble and lifelong learning,'' in \emph{International Conference
  on Learning Representations}, 2020. [Online]. Available:
  \url{https://openreview.net/forum?id=Sklf1yrYDr}
\BIBentrySTDinterwordspacing

\bibitem{Lorch2019}
B.~Lorch, S.~Agarwal, and H.~Farid, ``{Forensic Reconstruction of Severely
  Degraded License Plates},'' \emph{Electronic Imaging}, vol. 2019, no.~5,
  2019.

\bibitem{Schirrmacher2020}
F.~Schirrmacher, B.~Lorch, B.~Stimpel, T.~K{\"o}hler, and C.~Riess, ``Sr$^2$:
  Super-resolution with structure-aware reconstruction,'' in \emph{2020 IEEE
  International Conference on Image Processing (ICIP)}, 2020, p. 533537.

\bibitem{Rossi2021}
G.~Rossi, M.~Fontani, and S.~Milani, ``Neural network for denoising and reading
  degraded license plates,'' in \emph{Pattern Recognition. ICPR International
  Workshops and Challenges: Virtual Event, January 10--15, 2021, Proceedings,
  Part VI}.\hskip 1em plus 0.5em minus 0.4em\relax Springer International
  Publishing, 2021, pp. 484--499.

\bibitem{8462282}
M.~Zhang, W.~Liu, and H.~Ma, ``Joint license plate super-resolution and
  recognition in one multi-task gan framework,'' in \emph{2018 IEEE
  International Conference on Acoustics, Speech and Signal Processing
  (ICASSP)}, 2018, pp. 1443--1447.

\bibitem{Lee_2019_ICCV}
Y.~Lee, J.~Lee, H.~Ahn, and M.~Jeon, ``Snider: Single noisy image denoising and
  rectification for improving license plate recognition,'' in \emph{Proceedings
  of the IEEE/CVF International Conference on Computer Vision (ICCV)
  Workshops}, Oct 2019.

\bibitem{Caruana1993}
R.~Caruana, ``Multitask learning: A knowledge-based source of inductive bias,''
  in \emph{ICML}, 1993.

\bibitem{Caruana1997}
------, ``Multitask learning,'' \emph{Machine learning}, vol.~28, no.~1, pp.
  41--75, 1997.

\bibitem{Ruder2017}
S.~Ruder, ``An overview of multi-task learning in deep neural networks,''
  \emph{arXiv preprint arXiv:1706.05098}, 2017.

\bibitem{Villar2021}
A.~Villar-Corrales, F.~Schirrmacher, and C.~Riess, ``Deep learning
  architectural designs for super-resolution of noisy images,'' in \emph{ICASSP
  2021-2021 IEEE International Conference on Acoustics, Speech and Signal
  Processing (ICASSP)}.\hskip 1em plus 0.5em minus 0.4em\relax IEEE, 2021, pp.
  1635--1639.

\bibitem{Shashirangana2020}
J.~Shashirangana, H.~Padmasiri, D.~Meedeniya, and C.~Perera, ``Automated
  license plate recognition: a survey on methods and techniques,'' \emph{IEEE
  Access}, vol.~9, pp. 11\,203--11\,225, 2020.

\bibitem{Du2012}
S.~Du, M.~Ibrahim, M.~Shehata, and W.~Badawy, ``Automatic license plate
  recognition (alpr): A state-of-the-art review,'' \emph{IEEE Transactions on
  circuits and systems for video technology}, vol.~23, no.~2, pp. 311--325,
  2012.

\bibitem{Anagnostopoulos2008}
C.-N.~E. Anagnostopoulos, I.~E. Anagnostopoulos, I.~D. Psoroulas, V.~Loumos,
  and E.~Kayafas, ``License plate recognition from still images and video
  sequences: A survey,'' \emph{IEEE Transactions on intelligent transportation
  systems}, vol.~9, no.~3, pp. 377--391, 2008.

\bibitem{Wen2011}
Y.~Wen, Y.~Lu, J.~Yan, Z.~Zhou, K.~M. von Deneen, and P.~Shi, ``An algorithm
  for license plate recognition applied to intelligent transportation system,''
  \emph{IEEE Transactions on intelligent transportation systems}, vol.~12,
  no.~3, pp. 830--845, 2011.

\bibitem{Liu2010}
L.~Liu, H.~Zhang, A.~Feng, X.~Wan, and J.~Guo, ``Simplified local binary
  pattern descriptor for character recognition of vehicle license plate,'' in
  \emph{2010 Seventh International Conference on Computer Graphics, Imaging and
  Visualization}.\hskip 1em plus 0.5em minus 0.4em\relax IEEE, 2010, pp.
  157--161.

\bibitem{Redmon2016}
J.~Redmon, S.~Divvala, R.~Girshick, and A.~Farhadi, ``You only look once:
  Unified, real-time object detection,'' in \emph{Proceedings of the IEEE
  conference on computer vision and pattern recognition}, 2016, pp. 779--788.

\bibitem{Ren2016}
S.~Ren, K.~He, R.~Girshick, and J.~Sun, ``Faster r-cnn: towards real-time
  object detection with region proposal networks,'' \emph{IEEE transactions on
  pattern analysis and machine intelligence}, vol.~39, no.~6, pp. 1137--1149,
  2016.

\bibitem{Tan2020}
M.~Tan, R.~Pang, and Q.~V. Le, ``Efficientdet: Scalable and efficient object
  detection,'' in \emph{Proceedings of the IEEE/CVF conference on computer
  vision and pattern recognition}, 2020, pp. 10\,781--10\,790.

\bibitem{Krizhevsky2012}
A.~Krizhevsky, I.~Sutskever, and G.~E. Hinton, ``Imagenet classification with
  deep convolutional neural networks,'' \emph{Advances in neural information
  processing systems}, vol.~25, pp. 1097--1105, 2012.

\bibitem{He2016}
K.~He, X.~Zhang, S.~Ren, and J.~Sun, ``Deep residual learning for image
  recognition,'' in \emph{Proceedings of the IEEE conference on computer vision
  and pattern recognition}, 2016, pp. 770--778.

\bibitem{Tan2019}
M.~Tan and Q.~Le, ``Efficientnet: Rethinking model scaling for convolutional
  neural networks,'' in \emph{International Conference on Machine
  Learning}.\hskip 1em plus 0.5em minus 0.4em\relax PMLR, 2019, pp. 6105--6114.

\bibitem{Li2018}
H.~Li, P.~Wang, and C.~Shen, ``Toward end-to-end car license plate detection
  and recognition with deep neural networks,'' \emph{IEEE Transactions on
  Intelligent Transportation Systems}, vol.~20, no.~3, pp. 1126--1136, 2018.

\bibitem{Silva2021}
S.~M. Silva and C.~R. Jung, ``A flexible approach for automatic license plate
  recognition in unconstrained scenarios,'' \emph{IEEE Transactions on
  Intelligent Transportation Systems}, 2021.

\bibitem{Zhang2019}
H.~Zhang, F.~Sun, X.~Zhang, and L.~Zheng, ``{License Plate Recognition Model
  Based on CNN + LSTM + CTC},'' in \emph{International Conference of Pioneering
  Computer Scientists, Engineers and Educators}.\hskip 1em plus 0.5em minus
  0.4em\relax Springer, 2019, pp. 657--678.

\bibitem{He2017}
K.~He, G.~Gkioxari, P.~Doll{\'a}r, and R.~Girshick, ``Mask r-cnn,'' in
  \emph{Proceedings of the IEEE international conference on computer vision},
  2017, pp. 2961--2969.

\bibitem{Qiao2020}
L.~Qiao, Y.~Chen, Z.~Cheng, Y.~Xu, Y.~Niu, S.~Pu, and F.~Wu, ``Mango: A mask
  attention guided one-stage scene text spotter,'' \emph{arXiv preprint
  arXiv:2012.04350}, 2020.

\bibitem{Agarwal2017}
S.~Agarwal, D.~Tran, L.~Torresani, and H.~Farid, ``{Deciphering Severely
  Degraded License Plates},'' \emph{Electronic Imaging}, vol. 2017, no.~7, pp.
  138--143, 2017.

\bibitem{Shi2016}
B.~Shi, X.~Bai, and C.~Yao, ``{An End-to-End Trainable Neural Network for
  Image-based Sequence Recognition and Its Application to Scene Text
  Recognition},'' \emph{IEEE Transactions on Pattern Analysis and Machine
  Intelligence}, vol.~39, no.~11, pp. 2298--2304, 2016.

\bibitem{Shivakumara2018}
P.~Shivakumara, D.~Tang, M.~Asadzadehkaljahi, T.~Lu, U.~Pal, and M.~H. Anisi,
  ``{CNN-RNN based method for license plate recognition},'' \emph{CAAI
  Transactions on Intelligence Technology}, vol.~3, no.~3, pp. 169--175, 2018.

\bibitem{Suvarnam2019}
B.~Suvarnam and V.~S. Ch, ``{Combination of CNN-GRU Model to Recognize
  Characters of a License Plate number without Segmentation},'' in \emph{5th
  International Conference on Advanced Computing \& Communication Systems},
  2019, pp. 317--322.

\bibitem{Graves2006}
A.~Graves, S.~Fern{\'a}ndez, F.~Gomez, and J.~Schmidhuber, ``Connectionist
  temporal classification: labelling unsegmented sequence data with recurrent
  neural networks,'' in \emph{Proceedings of the 23rd international conference
  on Machine learning}, 2006, pp. 369--376.

\bibitem{Hilario2017}
H.~Seibel, S.~Goldenstein, and A.~Rocha, ``Eyes on the target: Super-resolution
  and license-plate recognition in low-quality surveillance videos,''
  \emph{IEEE Access}, vol.~5, pp. 20\,020--20\,035, 2017.

\bibitem{Henry2020}
C.~Henry, S.~Y. Ahn, and S.-W. Lee, ``Multinational license plate recognition
  using generalized character sequence detection,'' \emph{IEEE Access}, vol.~8,
  pp. 35\,185--35\,199, 2020.

\bibitem{Li2019}
H.~Li, P.~Wang, and C.~Shen, ``Toward end-to-end car license plate detection
  and recognition with deep neural networks,'' \emph{IEEE Transactions on
  Intelligent Transportation Systems}, vol.~20, no.~3, pp. 1126--1136, 2019.

\bibitem{DerKiureghian2009}
A.~Der~Kiureghian and O.~Ditlevsen, ``Aleatory or epistemic? does it matter?''
  \emph{Structural safety}, vol.~31, no.~2, pp. 105--112, 2009.

\bibitem{Hendrycks2017}
D.~Hendrycks and K.~Gimpel, ``A baseline for detecting misclassified and
  out-of-distribution examples in neural networks,'' \emph{Proceedings of
  International Conference on Learning Representations}, 2017.

\bibitem{Guo2017}
C.~Guo, G.~Pleiss, Y.~Sun, and K.~Q. Weinberger, ``On calibration of modern
  neural networks,'' in \emph{International Conference on Machine
  Learning}.\hskip 1em plus 0.5em minus 0.4em\relax PMLR, 2017, pp. 1321--1330.

\bibitem{Hinton1993}
G.~E. Hinton and D.~Van~Camp, ``Keeping the neural networks simple by
  minimizing the description length of the weights,'' in \emph{Proceedings of
  the sixth annual conference on Computational learning theory}, 1993, pp.
  5--13.

\bibitem{Graves2011}
A.~Graves, ``Practical variational inference for neural networks,'' in
  \emph{Advances in neural information processing systems}.\hskip 1em plus
  0.5em minus 0.4em\relax Citeseer, 2011, pp. 2348--2356.

\bibitem{Blundell2015}
C.~Blundell, J.~Cornebise, K.~Kavukcuoglu, and D.~Wierstra, ``Weight
  uncertainty in neural network,'' in \emph{International Conference on Machine
  Learning}.\hskip 1em plus 0.5em minus 0.4em\relax PMLR, 2015, pp. 1613--1622.

\bibitem{Ioffe2015}
S.~Ioffe and C.~Szegedy, ``Batch normalization: Accelerating deep network
  training by reducing internal covariate shift,'' in \emph{International
  conference on machine learning}.\hskip 1em plus 0.5em minus 0.4em\relax PMLR,
  2015, pp. 448--456.

\bibitem{Dong2016}
C.~Dong, C.~C. Loy, and X.~Tang, ``Accelerating the super-resolution
  convolutional neural network,'' in \emph{European conference on computer
  vision}.\hskip 1em plus 0.5em minus 0.4em\relax Springer, 2016, pp. 391--407.

\bibitem{Srivastava2014}
N.~Srivastava, G.~Hinton, A.~Krizhevsky, I.~Sutskever, and R.~Salakhutdinov,
  ``Dropout: a simple way to prevent neural networks from overfitting,''
  \emph{The journal of machine learning research}, vol.~15, no.~1, pp.
  1929--1958, 2014.

\bibitem{Fort2019}
S.~Fort, H.~Hu, and B.~Lakshminarayanan, ``Deep ensembles: A loss landscape
  perspective,'' \emph{arXiv preprint arXiv:1912.02757}, 2019.

\bibitem{Krogh1994}
A.~Krogh and J.~Vedelsby, ``Neural network ensembles, cross validation, and
  active learning,'' in \emph{Advances in Neural Information Processing Systems
  7, {[NIPS} Conference, Denver, Colorado, USA, 1994]}, G.~Tesauro, D.~S.
  Touretzky, and T.~K. Leen, Eds.\hskip 1em plus 0.5em minus 0.4em\relax {MIT}
  Press, 1994, pp. 231--238.

\bibitem{Xu2018}
Z.~Xu, W.~Yang, A.~Meng, N.~Lu, H.~Huang, C.~Ying, and L.~Huang, ``Towards
  end-to-end license plate detection and recognition: A large dataset and
  baseline,'' in \emph{Proceedings of the European conference on computer
  vision (ECCV)}, 2018, pp. 255--271.

\end{thebibliography}
